\documentclass[sigconf]{acmart}

\usepackage{enumitem}

\AtBeginDocument{%
  \providecommand\BibTeX{{%
    \normalfont B\kern-0.5em{\scshape i\kern-0.25em b}\kern-0.8em\TeX}}}

\copyrightyear{2022}
\acmYear{2022}
\setcopyright{acmlicensed}\acmConference[MM '22]{Proceedings of the 30th ACM International Conference on Multimedia}{October 10--14, 2022}{Lisboa, Portugal}
\acmBooktitle{Proceedings of the 30th ACM International Conference on Multimedia (MM '22), October 10--14, 2022, Lisboa, Portugal} \acmPrice{15.00}
\acmDOI{10.1145/3503161.3548080} \acmISBN{978-1-4503-9203-7/22/10}

\begin{document}

\title{Extreme-scale Talking-Face Video Upsampling with Audio-Visual Priors}

\author{Sindhu B Hegde}
\authornote{Both authors contributed equally to this research.}
\email{sindhu.hegde@research.iiit.ac.in}
\affiliation{%
  \institution{International Institute of Information Technology} \country{Hyderabad, India}
}

\author{Rudrabha Mukhopadhyay}
\email{radrabha.m@research.iiit.ac.in}
\authornotemark[1]
\affiliation{%
  \institution{International Institute of Information Technology}
  \country{Hyderabad, India}
}

\author{Vinay P Namboodiri}
\email{vpn22@bath.ac.uk}
\affiliation{%
  \institution{University of Bath} 
  \country{United Kingdom}
}

\author{C. V. Jawahar}
\email{jawahar@iiit.ac.in}
\affiliation{%
  \institution{International Institute of Information Technology} \country{Hyderabad, India}
}

\renewcommand{\shortauthors}{Sindhu B Hegde, Rudrabha Mukhopadhyay, Vinay P Namboodiri, \& C. V. Jawahar}
\begin{abstract}
  In this paper, we explore an interesting question of what can be obtained from an $8\times8$ pixel video sequence. Surprisingly, it turns out to be quite a lot. We show that when we process this $8\times8$ video with the right set of audio and image priors, we can obtain a full-length, $256\times256$ video. We achieve this $32\times$ scaling of an extremely low-resolution input using our novel audio-visual upsampling network. The audio prior helps to recover the elemental facial details and precise lip shapes and a single high-resolution target identity image prior provides us with rich appearance details. Our approach is an end-to-end multi-stage framework. The first stage produces a coarse intermediate output video that can be then used to animate single target identity image and generate realistic, accurate and high-quality outputs. Our approach is simple and performs exceedingly well (an $8\times$ improvement in FID score) compared to previous super-resolution methods. We also extend our model to talking-face video compression, and show that we obtain a $3.5\times$ improvement in terms of bits/pixel over the previous state-of-the-art. The results from our network are thoroughly analyzed through extensive ablation experiments (in the paper and supplementary material). We also provide the demo video along with code and models on our website\footnote{\url{http://cvit.iiit.ac.in/research/projects/cvit-projects/talking-face-video-upsampling}}.
\end{abstract}

\ccsdesc[500]{Computing methodologies~Reconstruction}
\ccsdesc[500]{Computing methodologies~Neural networks}

\keywords{video upsampling, video super-resolution, video compression, talking-face videos, audio-visual learning}

\begin{teaserfigure}
  \includegraphics[width=\textwidth]{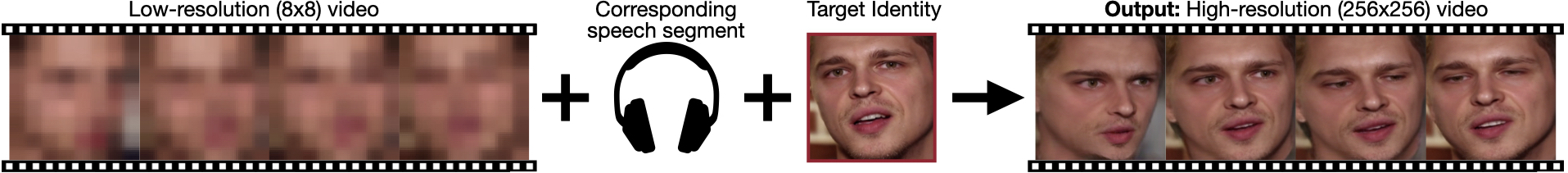}
  \vspace{-20pt}
  \caption{We solve the problem of upsampling extremely low-resolution (LR) talking-face videos to generate high-resolution (HR) outputs. Our approach exploits LR frames ($8\times8$ pixels), corresponding audio signal and a single HR target identity image to synthesize realistic, high-quality talking-face videos ($256\times256$ pixels). Please check our project page for video results.}
  \label{fig:teaser}
\end{teaserfigure}

\maketitle

\vspace{-10pt}
\section{Introduction}
\label{section:introduction}

Over the years, we have always been fascinated with questions that can push the limits of computer vision. For instance, can we recognize actions~\cite{efros} or objects~\cite{torralba_2006} in videos/images of about $30\times30$ pixels? Or how small a face can we detect in an image? It turns out, the face can be as tiny as $3$ pixels tall~\cite{Hu_2017_CVPR}! Recovering extremely feeble signals has also led to remarkable achievements, such as the imaging of the black hole. In this vein, we explore whether we can upsample talking-face videos with resolutions as low as $8\times8$ pixels. Clearly, this is an exceptionally challenging task. Surprisingly, we reveal that one can obtain realistic high-resolution (HR) talking-faces ($256\times256$ pixels) when provided with the right set of additional information. We utilize a single target identity image and the accompanying audio to upsample $8\times8$ video to a full $256\times256$ dimensional video (a fascinating $32\times$ scale-factor) that far exceeds previous methods. In today's digitally connected world, where talking-face videos are among the most common forms of video content, our multi-modal system can have numerous potential applications. Some of them include: (i) video conferencing in low-bandwidth situations, (ii) recovering low-quality archival footage of public talks and speeches and (iii) enhancing videos captured from a distance with a high camera zoom.

\vspace{5pt}
\noindent
\textbf{Challenges:}
Although our task has promising applications, synthesizing HR videos from extremely low-resolution (LR) inputs (e.g., $8\times8$ pixels) is a very challenging task. Essential face attributes such as identity, age and gender are almost entirely lost at such low resolutions and cannot be directly recovered (see LR video in Figure~\ref{fig:teaser}). Apart from reconstructing these elementary identity details, the network must also learn to predict the original head poses and accurate lip shapes. Given the arduous nature of the task, it is evident that the model will struggle to achieve the desired quality if it relies solely on the LR input. Thus, in our work, we argue that considering appropriate priors is quintessential to obtain high-quality results. Specifically, we assess the importance of two kinds of priors: (i) audio signal, which can help to recover elemental facial attributes and can significantly improve lip shape generation; (ii) a single HR target identity image which can aid in restoring fine-grained details such as skin texture, color, hair, teeth and surrounding background. The target identity image can be any sample frame, either from the same video or any other image with similar characteristics in terms of the face identity, pose, clothing and background. To understand our task better, we will now explore how our work is connected to some of the existing problems in literature. 

\vspace{5pt}
\noindent
\textbf{Super-Resolution (SR) Perspective:}
The SR literature till date has focused on super-resolving inputs (either faces, generic images or videos)~\cite{progressive-face-sr, dbpn, RBPN2019, tecogan} where sufficient information is already available (e.g., $256\times256$ pixels input). None of these methods can handle extreme scale-factors like $32\times$. When the input resolution is very low (like $8\times8$), we observe that most of the current SR works~\cite{progressive-face-sr, dicgan, dbpn, SPARNet, tecogan, frvsr, RBPN2019} generate sub-optimal results. The essential visual attributes such as the identity, face texture and lip shape do not accurately match the original face. This is natural because the network is forced to speculate these details without adequate priors. Thus, the existing methods: (i) do not aim to preserve the specific identity details and (ii) do not explicitly deal with talking-face videos where specialized temporal information like lip synchronisation must be maintained throughout the video.

In our work, we propose to address these limitations by generating high-quality talking-face videos with accurate lip-sync. It is important to note that although we aim to generate HR talking-faces from LR inputs, the task we are attempting is very different compared to the typical SR problem. The use of a single HD target identity image and synthesizing from extremely LR talking-faces ($8\times8$ pixels) sets us considerably apart from the traditional SR.

\vspace{5pt}
\noindent
\textbf{Compression Perspective:}
Our task enables applications such as low-bandwidth video calling; thus making it closely related to the task of talking-face video compression. However, unlike the existing works like ``os-synth''~\cite{nvidia_compression_2021_cvpr} where 3D face keypoints are transmitted, we take a unique path in our work. We propose to transmit the LR frames to extract the face structure, motion and pose information. Transmitting keypoints have several limitations: (i) Keypoints can only be extracted if we have the HR video beforehand. While the availability of actual video might be a possibility in video conferencing, this poses severe constraints in various other applications where the actual HD video is not present; (ii) Keypoints do not encode adequate head pose information, thus requiring additional specialized head pose estimation models; (iii) Keypoints do not cover other information like background, lighting, accessories like glasses and beards that could be present in talking-faces. Thus, in our work, we demonstrate the advantages of using LR frames and achieve better compression ratio compared to the standard codecs, while also not compromising on the desired quality.

\vspace{5pt}
\noindent
\textbf{Talking-Face Animation Perspective:}
Our task also shares similarities with audio-driven talking-face generation (A2TF)~\cite{YST_ijcv_2019, lipgan_2019, wav2lip, Yang:2020:MakeItTalk, talkingfaceacmmm} and face re-enactment (FR)~\cite{fomm_nips_2019, PC_AVS_2021_cvpr} tasks. A2TF works aim to generate videos of a target identity conditioned on the audio signal (input: single target identity image + audio). FR methods ingest a single target identity image and an HD video of a different identity with an aim to animate the target image according to the motion of the driving video (input: single target identity image + HD video as pose prior). Although we agree that in terms of the problem space, our work resembles A2TF/FR, we want to point out that our focus is very different - upsampling extremely low-resolution videos. There are key differences as noted in Table~\ref{table:tfgen_differences}. We leverage positive aspects from these dimensions to solve an entirely new task - generating HD videos from extremely LR inputs while preserving the exact same facial features, e.g., pose and identity. Nevertheless, we also include a comparison with A2TF and FR works in Section~\ref{section:A2TH_comparison}.

\vspace{-12pt}
\begin{table}[ht]
    \centering
    \small
    \setlength{\tabcolsep}{3pt}
    \caption{Key differences between audio-driven talking-face generation (A2TF), face re-enactment (FR) and proposed task.}

    \vspace{-8pt}
    \begin{tabular}{l|c|c|c|c}
    \hline

    & Pose prior & Lips are in-sync & Matches GT & Can be used \\
    & used & with audio? & frames? & for SR?\\
     
    \hline
    
    A2TF & None & $\checkmark$ & $\times$ (changes pose) & $\times$\\
    
    FR & HD video & $\checkmark$ & $\checkmark$ (same id-recons.) & $\times$\\
    
    \hline
    
    \textbf{Ours} & LR frames & \textbf{$\checkmark$} & \textbf{$\checkmark$} & \textbf{$\checkmark$}\\

    \hline
    
    \end{tabular}
    \vspace{-10pt}
    \label{table:tfgen_differences}
\end{table}

\begin{figure}[t]
  \includegraphics[width=\linewidth]{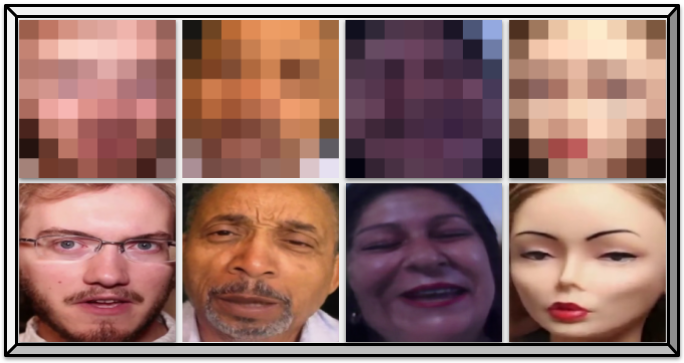}
  \vspace{-20pt}
  \caption{Input: LR frames (top); Output: HR frames (bottom). Our approach effectively synthesizes a diverse set of HR talking-faces, handling different poses, gender, age and race. Our method also works for ``human-like'' synthetic faces even though it was trained only on real faces.}
  \label{fig:gallery}
  \vspace{-15pt}
\end{figure}

\vspace{-5pt}
\noindent
\textbf{Overview of this Work:}
In this work, we propose a talking-face video upsampling framework, where the core idea is to utilize adequate priors to generate high-quality ($256\times256$) videos from extremely low-resolution inputs. A gallery showing our synthesized frames is displayed in Figure~\ref{fig:gallery}. We conduct extensive experiments and comparisons with state-of-the-art methods for the tasks of super-resolution and compression. To the best of our knowledge, we are the first to synthesize high-quality talking-faces at scale-factors of $32\times$ from an input as small as $8\times8$ pixels. We also show how our network can be utilized for low-bandwidth video conferencing, along with a demo video on our project page. 

To summarize, the major contributions of our work are:

\begin{itemize}[topsep=0.8pt, noitemsep, leftmargin=3mm]

    \item We present a novel audio-visual network that can upsample very low-resolution talking-face videos at scale-factors previously unseen in video SR literature ($32\times$). 
    
    \item Our approach to make use of adequate priors: (i) audio signal and (ii) a single target identity, achieves significant improvements over the existing works for SR and compression tasks.
    
    \item Our system has strong real-world use-cases, owing to its ability to preserve true identity information. We specifically demonstrate the superior quality of our results for low-bandwidth video conferencing and achieve a significant reduction in bandwidth compared to H.264 standard.
    
\end{itemize}

\section{Related Work}
\label{section:related_works}

\noindent
\textbf{Image \& Video Super-Resolution:}
Single image super-resolution (SISR) has progressed tremendously with the advent of deep learning and CNNs~\cite{dong_first_sr}. A flurry of works~\cite{vdsr, edsr, ntire_report, dbpn, talkingfaceacmmm, depthSR, SRusingdual} have followed this initial method, thereby improving the performance by many folds. To account for specialized facial attributes which are not present in generic images, face super-resolution (FSR) methods~\cite{SPARNet, face_sr_somework, KIM202111} with specific loss functions like facial landmarks~\cite{dicgan} and facial heatmaps~\cite{progressive-face-sr} were proposed. However, these FSR approaches focus on scale-factors up to $8\times$, which is in stark contrast to our attempt of $32\times$ SR.  

Recently, PULSE~\cite{PULSE_CVPR_2020} proposed a StyleGAN-based approach to super-resolve static faces at high scale-factors of $64\times$. However, the model generates imaginary faces of people who do not exist, posing severe constraints on applications where the person's identity needs to be matched/recovered correctly. Conversely, in our work, we aim to generate HR faces of the specific (real) identity.

To capture the temporal aspect which is not present in the static faces, video SR approaches came into picture. From early recurrent architectures~\cite{video_sr_1} to more recent advancements~\cite{frvsr, RBPN2019, tecogan}, impressive results have been achieved for scale-factors up to $4\times$. However, these generic video SR methods produce blurred results with many artifacts when the input resolution is very low ($8\times8$) and the scale-factors are very high ($>8\times$). Moreover, ``talking-face videos'' have their own set of additional challenges, which are neither tackled in video SR nor in static FSR works. Thus, in our work, we overcome these limitations and design a novel network that specifically deals with talking-face videos.

\vspace{5pt}
\noindent
\textbf{Talking-Face Video Generation:}
Audio-driven talking-face generation (A2TF) is an active research area. Various methods have been proposed~\cite{text-based-editing, Yao2020IterativeTE, wav2lip, talking_face_video_sig, Yang:2020:MakeItTalk} to accurately morph the lip movements of input target identity to be in-sync with the corresponding speech. Face re-enactment (FR) works~\cite{face2face, deep_video_portraits, fomm_nips_2019} have also shown impressive performance in transferring head motions and expressions based on guiding videos. While our task shares some similarities with A2TF/FR works, we differ in the fact that we are given very sparse information in the form of very low-resolution input. This is not the case in talking-face generation models where HR frames (or extracted landmarks) are used for conditional generation.

Few works~\cite{deep_video_portraits, facial_iccv_2021} use a 3D model as an intermediate step to recover high-quality videos. The major benefit of adapting a 3D model is the superior quality of the output generations. But, it comes with an additional overhead - the computational complexity, making such heavy models very impractical for mobile hardware deployments. In contrast, our method is (i) simple, (ii) generalised, since it can be applied for any in-the-wild identity (unlike some of the 3D models that require speaker-specific training) and (iii) does not require specialized large-scale 3D datasets to train the models.

\vspace{5pt}
\noindent
\textbf{Data Compression using Deep Learning:}
Deep learning has lead to profound improvements in image and video compression techniques. Starting from initial auto-encoder based methods~\cite{CompressionAE_icassp_2018, CompressionNN_nips_2014} to flow-based approaches~\cite{Agustsson_2020_CVPR, CompressionFlow_nips_2019}, there have been multiple efforts to obtain a compact image/video representation. To enable video calls at a reduced bandwidth, specific methods like~\cite{VideoConfMedical_2013_atnac, VideoChatCompression_2021_cvprw} have been designed. In ``SRVC''~\cite{SRVC_2021_ICCV} authors used video SR as a tool for compression, but at a scale-factor of merely $2\times$. Recently, ``os-synth''~\cite{nvidia_compression_2021_cvpr} demonstrated impressive results by transmitting a sequence of learned 3D facial keypoints. However, as discussed previously, transmitting keypoints has its own set of limitations. Thus, unlike the existing compression methods, we take a path of extreme-scale ($32\times$) SR for the first time and transmit very low-resolution ($8\times8$) videos. We show that our proposed approach of utilizing the audio and the visual modalities can enable video-conferencing in bandwidth-limited regions, while also achieving a better compression ratio over the existing works.

\section{Learning to Upsample LR Videos}
\label{section:methodology}

We start the discussion by highlighting the critical elements of our approach. We then present our framework with a detailed description of the modules involved.

\vspace{-5pt}
\subsection{Critical Elements of our Approach}

\noindent
\textbf{Audio Prior:}
\label{issue:ambiguous_nature}
As discussed previously, when the input resolution is a meager $8\times8$ pixel video, the ambiguity and the loss of information is so paramount that the person's original identity characteristics are barely discernible. In such situations, we show that audio can aid in the recovery of dominant facial traits of the person because the audio and the face share multiple common features~\cite{speech2face, SeeingVoices_cvpr_2018, FaceVoiceAssociations_accv_2018} like gender, age and ethnicity. We exploit the audio signal not only to disambiguate the LR input, but also to greatly improve the lip shape generation. Although precise lip shape is not a crucial necessity for static face SR, it is a very important aspect of video SR where the generated lip movements should sync with the given speech. In our work, we explore the natural correlation between speech and lips~\cite{wav2lip, conversation, Hegde_2021_WACV, perfect_match} to generate accurate lip movements. \\ 

\vspace{-5pt}
\noindent
\textbf{Visual Prior:}
\label{issue:inadequate_prior}
To be able to generate faces that replicate the actual identity, it is important to preserve sharp details like face texture, lip colour, hair, teeth and skin tone. Most of the current works hallucinate these details, leading to significant variations in fine-grained information. Such models are thus unusable for real-world use-cases, where videos of a specific identity needs to be generated. We argue that considering the adequate prior information is of utmost importance to: (i) generate a video of a specific identity and (ii) reconstruct the high-quality facial details. To achieve this, we provide our network with a single HR image of the target identity, which helps to transfer the identity-specific sharp details to the synthesized video. In most applications, a single HR identity image is easily accessible. For example, during video conferencing, the first frame can be transmitted in the original resolution.

\vspace{-5pt}
\subsection{Our Approach}

\begin{figure*}[t]
  \includegraphics[width=\textwidth]{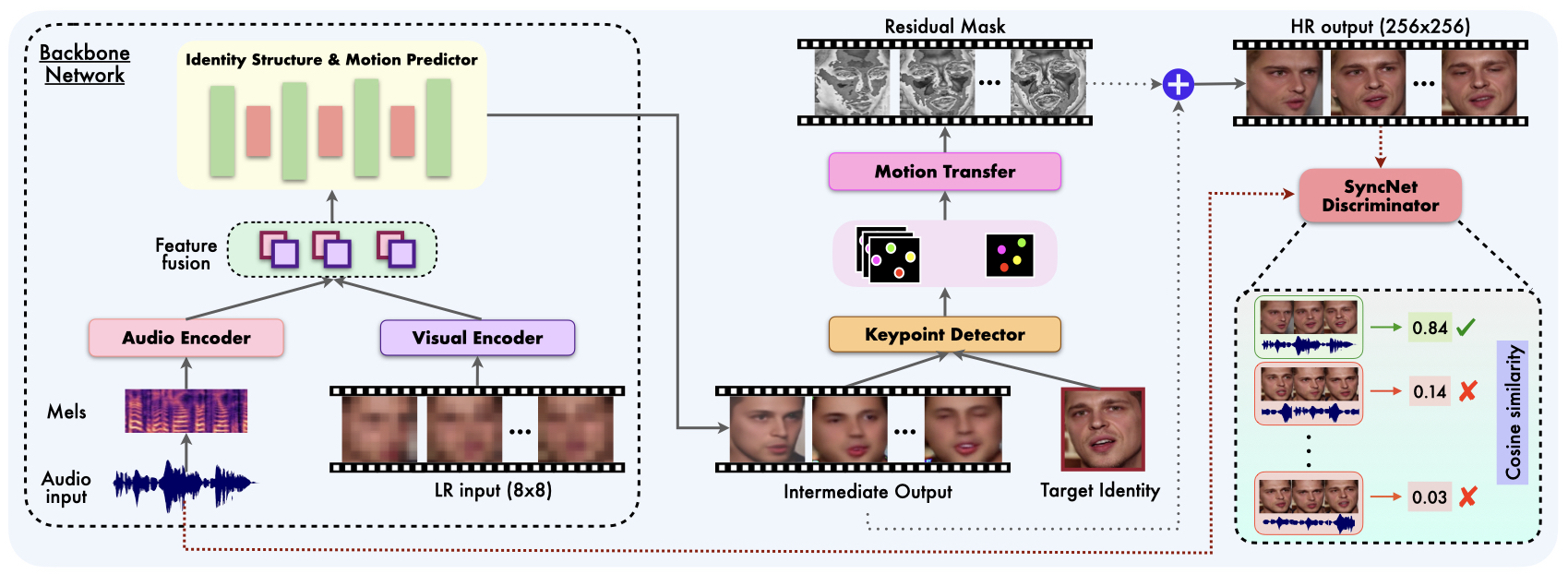}
  \vspace{-20pt}
  \caption{We propose a novel audio-visual talking-face upsampling framework. Our approach recovers the basic facial attributes like the identity structure, pose and motion using a backbone network. We then consider a single HR target identity image to capture the fine-grained details. Our end-to-end trainable animation network ingests the intermediate outputs (from backbone network) and the target identity and reconstructs realistic, high-quality videos that accurately match the actual identity.}
  \label{fig:architecture}
  \vspace{-10pt}
\end{figure*}

An overview of our proposed framework is depicted in Figure~\ref{fig:architecture}. The goal of our work is to generate a sequence of HR frames, $F_{hr} = (hr_1, hr_2, ..., hr_n)$ from the LR input, $F_{lr} = (lr_1, lr_2, ..., lr_n)$ that accurately match the ground-truth frames $F_{gt} = (gt_1, gt_2, ..., gt_n)$. We consider the corresponding audio signal, $A$ and a single HR target identity, $F_{id}$ as prior information. We detail the different components of our model and the network architecture below. 

\vspace{3pt}
\noindent
\textbf{Backbone Network:}
We pre-train a backbone network to generate a ``driving video'' for our face animation network. It extracts the relevant features of the identity, face structure, pose and motion information. We observe that pre-training the network before training the face animation network helps to improve the model's performance because the basic face details and motion information captured is essential to animate the HR face (details about the animation network is explained below). We empirically demonstrate the importance of this in the ablation study (in supplementary).

\noindent
\underline{\textit{Visual Encoder:}} We extract the visual features from the LR frames $F_{lr}$ using a visual encoder, which comprises a series of $3$D convolution layers with residual connections. Our visual encoder resembles multiple previous models~\cite{avobjects, Hegde_2021_WACV, perfect_match} designed specifically for processing talking-face videos. The input to the visual encoder is a contiguous window of $T$ LR frames $l: (N, T, 3, 8, 8)$ where $N$ refers to the batch size and $T$ refers to the window frames (here $T$=5). The encoder processes these input frames and generates the visual embeddings, $f_v: (N, T, 512, 8, 8)$.

\noindent
\underline{\textit{Audio Encoder:}} We consider the corresponding audio segment $A$ and extract the melspectrogram representation using a window length of $25$ms with a hop length of $10$ms sampled at $16$kHz. The melspectrograms $(T', 80)$ are given to the audio encoder, which is a stack of residual $1$D convolutions with appropriate strides to match the visual time-steps $T$. The generated features ($N, T, 512$) are then upsampled using transpose convolution layers to obtain the audio embeddings, $f_a: (N, T, 512, 8, 8)$. 

\noindent
\underline{\textit{Identity Structure $\&$ Motion Predictor:}} We concatenate the learned visual and audio embeddings in the latent space (along the channel dimension) to obtain $f_{cat} = (N, T, 1024, 8, 8)$. Inspired from Deep Back-Projection Network (DBPN)~\cite{dbpn}, we use iterative upsampling and downsampling layers in our module, where the primary idea is to effectively capture the mutual relationship between the LR and HR frames. We consider the fused representation $f_{cat}$ and stack the time steps along the batch dimension to obtain $f_{cat}: (N*T, 1024, 8, 8)$. As the visual and the audio encoders have already captured the temporal information, the stacking strategy improves the convergence speed and also gives us the desired performance. The output of this block is a sequence of frames ($F_{int}$) of resolution $256\times256$ pixels, which encapsulates the elemental facial details, face structure, pose and motion information. The network is trained to minimize the $L_1$ reconstruction loss:

\vspace{-5pt}   
\begin{equation}
L_{\mathrm{rec}} = \frac{1}{N} \sum^{N}_{i=1} ||F_{int} - F_{gt}||_1
\label{eqn:recon}
\end{equation}

\noindent
\textbf{Face Animation Network:}
To synthesize the HR videos of the target identity by preserving all the details, we consider a single HR image of the target identity as our input. This is in-line with the audio-driven talking-face generation works~\cite{wav2lip, PC_AVS_2021_cvpr, fomm_nips_2019, nvidia_compression_2021_cvpr} where a single target identity is considered to replicate the identity-specific details. The target identity image helps to capture the fine-grained features like face texture, skin tone, hair and lip color, which are otherwise not recovered in the existing FSR works~\cite{dicgan, dbpn, SPARNet}. These details are crucial, especially when our input is a mere $8\times8$ pixel video and make our model applicable in cases where we must match the actual identity to the maximum extent. We adopt one of the popular methods, FOMM~\cite{fomm_nips_2019} to animate the target identity based on the driving video obtained from our backbone network.

\noindent
\underline{\textit{Overview of First-Order Motion Model (FOMM):}} FOMM ingests a target identity image (to extract the appearance) and a driving video (to extract the pose and motion). The learned latent representation of motion in the driving video is combined with the target identity to synthesize the output video. During training, the model observes the target-driving image pairs and predicts a dense motion field which is later encoded using a keypoint detection network. The target image is then rendered according to the learned trajectories in the driving video. We refer the reader to~\cite{fomm_nips_2019} for more details about FOMM. Note that FOMM is designed to work even when the target image and the driving video are of different identities. However, this feature is not necessary in our work where the aim is to preserve the target identity, so we make appropriate modifications, as described below.

\noindent
\underline{\textit{Adapting FOMM to our Task:}} For our task at hand, the goal is to animate the HR target identity image $F_{id}$ in accordance with the motion of the driving video $F_{int}$. As described previously, our backbone network reconstructs the basic identity attributes like face structure, age and gender. We thus generate a residual mask as the output of the animation network and add the intermediate outputs $F_{int}$ (see Figure~\ref{fig:architecture}) to obtain a realistic HR talking-face video $F_{hr}$ of the target identity as the final output. We fine-tune the entire network (including the backbone network) end-to-end, by optimizing the FOMM loss $L_{fomm}$ and our task-specific losses, $L_{rec}$, $L_{region}$ and $L_{sync}$. The FOMM loss $L_{fomm}$ consists of: (i) a VGG-19 based perceptual loss at multiple resolutions and (ii) an equivariance constraint to enforce the model to predict consistent keypoints to known geometric transformations. We describe other task-specific losses that we use in out network below.

\noindent
\underline{\textit{Enforcing Local Correspondence:}} In our experiments, we observed that the model at times generates frames where the facial regions like eyes, eyebrows and lips are slightly off-position. This occurs if the target identity ingested by our animation network is significantly different from the driving video. Hence, to further improve the generation quality, we add a face landmark-based region loss, which penalizes the network for generating incorrect regions. We compute the face landmarks~\cite{fan_landmark_iccv_2017} for both GT $F_{gt}$ and generated frames $F_{hr}$ and extract the following $R$ face regions: lips, nose, eyes and eyebrows. We then add a patch-based local loss by minimizing the $L_2$ distance for all these $R$ regions (here $R$=4) to ensure that the predicted regions are as close as possible to the actual regions. 

\vspace{-10pt}
\begin{equation}
    L_\mathrm{region} = \frac{1}{N} \sum^{N}_{i=1}\sum^{R}_{r=1} ||F_{hr} - F_{gt}||_2
\label{eq:l2_region}
\end{equation}

\noindent
\underline{\textit{Enforcing Accurate Lip-sync:}} Prior works~\cite{wav2lip} involving speech and lip movements have shown that using a lip-sync discriminator can greatly benefit in enforcing strong audio-visual correspondences. This can also be observed in our task, as learning to synthesize HR frames from very low-resolution input might lead to generating lips that are out-of-sync with the audio segment. Thus, we pre-train a lip-sync discriminator, ``SyncNet'' adapted from~\cite{perfect_match}, trained to maximize the cosine similarity between the lip-speech pair $(F_{gt}, A)$ when they are in-sync (and minimise the similarity if they are out-of-sync). Once trained, we use this network as a frozen discriminator to penalize the generated frames $F_{hr}$ if they do not match the corresponding audio segment. In our end-to-end network, we minimize the sync loss:

\vspace{-10pt}
\begin{equation}
    L_\mathrm{sync} = - \frac{1}{N} \sum^{N}_{i=1} \log(\dfrac{f_{hr} \cdot a}{max(\lVert f_{hr} \rVert_{2} \cdot \lVert a \rVert_{2}, \epsilon)})
\label{eq:syncloss}
\end{equation}

\subsection{Training Settings and Datasets}

The final loss function is the combination of the above losses:

\vspace{-15pt}
\begin{equation}
    \resizebox{.9\hsize}{!}{$L_\mathrm{HR} = \lambda_{rec}L_\mathrm{rec} + L_{fomm} +  \lambda_{region}L_\mathrm{region} + \lambda_{sync}L_\mathrm{sync}$}
\label{eq:final_loss}
\end{equation}

\noindent
In our experiments, we set $\lambda_{rec}=50$, $\lambda_{region}=100$ and $\lambda_{sync}=0.05$. We provide the details regarding pre-processing and training settings in supplementary file on our project page.

\vspace{3pt}
\noindent
\textbf{Datasets:} 
We train our model using AVSpeech~\cite{avspeech_2018_tog} and VoxCeleb2~\cite{voxceleb2_2018_interspeech} datasets; both containing talking-face videos spanning a wide variety of identities, languages and poses. For AVSpeech data, we extract the face tracks using an off-the-shelf face detector~\cite{S3FDFaceDet_iccv_2017}. We curate a set of $50$ hours for training and $\sim3$ hours from the official test split for testing and verified it for accurate lip-sync using SyncNet~\cite{syncnet_2016}. We also benchmark our model on VoxCeleb2 data which comprises face tracks with a fair amount of background. Owing to computational limitations, we randomly sample a subset of $100$ hours for training and use the full official test split for testing. Note that there are no overlaps between the identities used in training and testing sets in both datasets. The code, models and file-lists are released on our website for reproducibility and future research.

\vspace{-3pt}
\section{Experiments}
\label{sec:results}

\subsection{Extreme-scale Super-Resolution}

\begin{table*}[ht]
    \centering
    \setlength{\tabcolsep}{8pt}
    \caption{Quantitative comparison for $32\times$ SR on AVSpeech~\cite{avspeech_2018_tog} and VoxCeleb2~\cite{voxceleb2_2018_interspeech} datasets. Our method outperforms the baselines by a significant margin across all metrics. Note that the baselines have also been trained with a single identity image.}

    \vspace{-10pt}
    \begin{tabular}{l|ccccc||ccccc}
    \hline

    \textbf{Dataset} & \multicolumn{5}{c||}{\textbf{AVSpeech}~\cite{avspeech_2018_tog}} & \multicolumn{5}{c}{\textbf{VoxCeleb2}~\cite{voxceleb2_2018_interspeech}}   \\
    \hline
    
    \textbf{Method} & 
    \textbf{PSNR}$\uparrow$ & \textbf{SSIM}$\uparrow$ & \textbf{FID}$\downarrow$ & \textbf{LMD}$\downarrow$ & \textbf{LSE-D}$\downarrow$ & 
    \textbf{PSNR}$\uparrow$ & \textbf{SSIM}$\uparrow$ & \textbf{FID}$\downarrow$ & \textbf{LMD}$\downarrow$ & \textbf{LSE-D}$\downarrow$ \\
     
    \hline
    
    Bicubic & 22.33 & 0.60 & 102.41 & 0.246 & 14.18 & 22.16 & 0.60 & 105.14 & 0.255 & 17.83\\
    
    SPARNet~\cite{SPARNet} & 23.17 & 0.68 & 92.14 & 0.201 & 12.87 & 22.98 & 0.67 & 83.01 & 0.228 & 14.07\\  
    
    TecoGAN~\cite{tecogan} & 19.26 & 0.62 & 84.73 & 0.213 & 13.01 & 16.91 & 0.54 & 82.19 & 0.234 & 14.12\\    

    \hline
    
    \textbf{Ours} & \textbf{25.06} & \textbf{0.73} & \textbf{11.54} & \textbf{0.162} & \textbf{12.43} & \textbf{24.95} & \textbf{0.71} & \textbf{14.10} & \textbf{0.196} & \textbf{13.91}\\
    
    \hline
    
    \end{tabular}
    \vspace{-5pt}
    \label{table:metrics}
\end{table*}

\begin{figure*}[ht]
  \includegraphics[width=\textwidth]{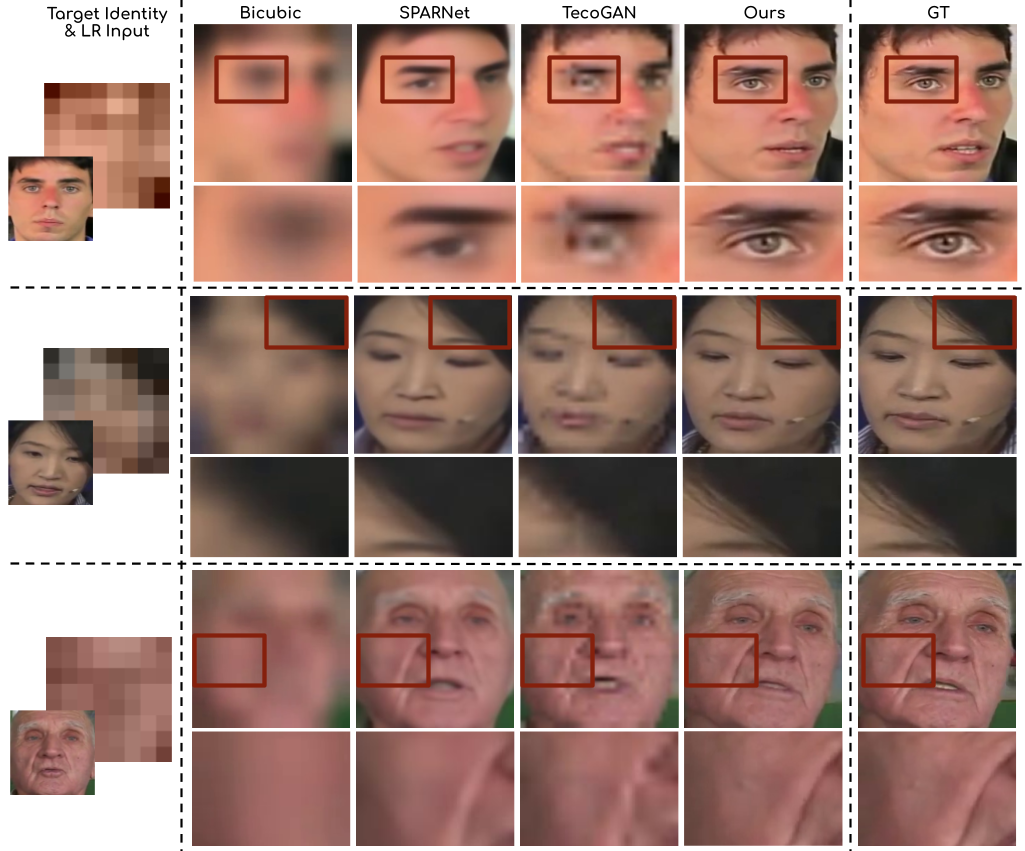}
  \vspace{-20pt}
  \caption{Qualitative comparisons on AVSpeech dataset~\cite{avspeech_2018_tog}. Our method  captures the rich identity-specific attributes like eyeballs, hair strands, face texture and lip shape, far better compared to the existing approaches.}
  \label{fig:comparison}
  \vspace{-10pt}
\end{figure*}

\begin{figure*}[ht]
  \includegraphics[width=\textwidth]{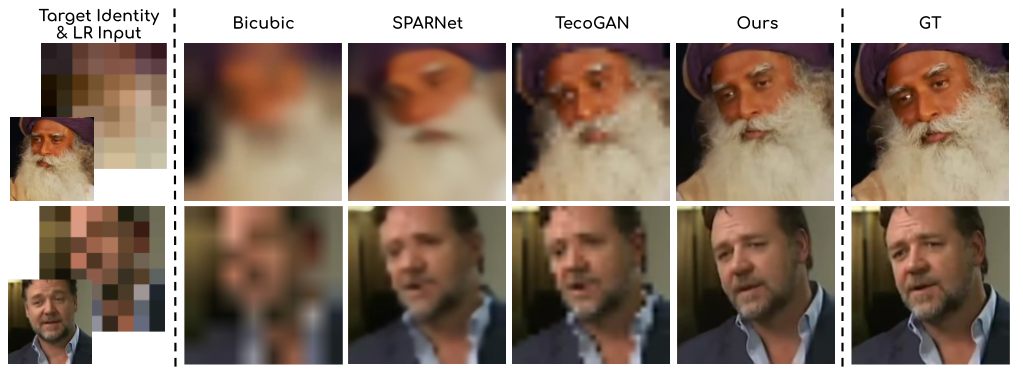}
  \vspace{-20pt}
  \caption{Qualitative comparisons on the models trained on VoxCeleb2 dataset~\cite{voxceleb2_2018_interspeech}. Our method surpasses the existing baselines in generating the outputs that accurately match the ground-truth identity.}
  \label{fig:comparison_vox2}
  \vspace{-10pt}
\end{figure*}

\noindent
\textbf{Baselines:}
The state-of-the-art works in video SR literature super-resolve up to a scale-factor of $4\times$. We thus re-train the existing state-of-the-art video SR method, TecoGAN~\cite{tecogan} at a scale-factor of $32\times$ on the same training dataset as ours. We extend the existing face SR approach, SPARNet~\cite{SPARNet} to work for video SR by appropriately modifying the architecture (ingest a window of $5$ frames) and train using the same settings as mentioned above. These methods originally do not consider a HR target identity as input; thus, it would be unfair to compare them without the identity information. Hence we provide a HR target identity image to these models in a manner typically used in talking-face generation methods: concatenating it channel-wise with the input.

\noindent
\textbf{Metrics:} 
We evaluate our SR model on: (i) PSNR, (ii) SSIM, (iii) Fréchet Inception Distance (FID)~\cite{heusel2018gans}, (iv) Landmark Distance (LMD) \cite{fan_landmark_iccv_2017} and (v) Lip-Sync Error Distance (LSE-D)~\cite{wav2lip}. More details about the metrics can be found in the supplementary file.

\noindent
\textbf{Results:} 
We compare our results with existing SR approaches at extreme scale-factor of $32\times$ in Table~\ref{table:metrics}. As we can see from the table, our method outperforms the existing works by a significant margin on both AVSpeech~\cite{avspeech_2018_tog} and VoxCeleb2~\cite{voxceleb2_2018_interspeech} datasets. None of the current techniques match the ground-truth identity (measured using PSNR) and perceptual quality (measured using FID) of our generations. The LSE-D metric indicates that our method achieves accurate lip-synchronization with audio, thus validating our claim that the audio signal enables us to generate far more accurate lip shapes than the competing methods. Our method also surpasses the existing approaches in preserving the overall face structure (measured using SSIM and LMD). 

Figures~\ref{fig:comparison} and~\ref{fig:comparison_vox2} show the qualitative comparisons. We can clearly observe that our models generate results with far fewer artifacts and captures rich, fine-grained details. Although all the comparison methods consider the HR target identity as input, they do not match the quality of our generations. This shows that our overall network design is highly effective in making use of the available target identity image. In the examples, we can also see the diverse range of our models' generative capabilities: eyeballs with precise eye color (Fig.~\ref{fig:comparison}: row $1$), microphone (Fig.~\ref{fig:comparison}: row $2$), hair strands (Fig.~\ref{fig:comparison}: row $2$), lip shape (Fig.~\ref{fig:comparison}: rows $2$,$3$ and Fig.~\ref{fig:comparison_vox2}: rows $1$,$2$), face texture such as wrinkles (Fig.~\ref{fig:comparison}: row $3$), beard (Fig.~\ref{fig:comparison_vox2}: rows $1$,$2$). More visual examples can be found on our project page.

\vspace{3pt}
\noindent
\textbf{Ablation study:}
We validate the design choices of our network by analyzing the importance of audio signal, landmark-based region loss, the use of different target identity images and several other additional experiments, along with human evaluations in our supplementary. We also compare the performance of different models at multiple scale-factors like $4\times$, $8\times$, $16\times$ and $32\times$ in supplementary.

\subsection{Talking-Face Video Compression}

\begin{figure}[t]
  \includegraphics[width=\linewidth]{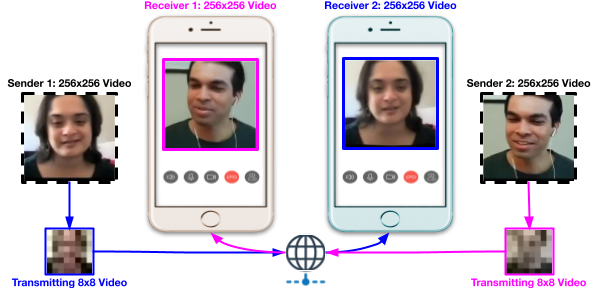}
  \vspace{-15pt}
  \caption{Illustration of low-bandwidth video calling enabled by our system. Note that HR frames shown at both the sender's end are taken from an actual video call recording (credits: \url{https://www.youtube.com/watch?v=lQJD8RAq3lY}). } 
  \label{fig:vidcall_demo}
  \vspace{-20pt}
\end{figure}

One of the major applications of our system is in compressing talking-face videos to reduce the bandwidth in video conferencing applications. We can transmit the LR frames ($8\times8$ pixels) along with the audio signal on the sender's side and the receiver can reconstruct the high-quality video ($256\times256$ pixels) using a single HR target identity image. A sample video calling demo is illustrated in Figure~\ref{fig:vidcall_demo}. We assume that a single target identity image can be sent at the beginning (e.g., $1^{st}$ frame) and hence does not consume additional bandwidth. Note that this is very different from the standard codecs, where full resolution I-frames are transmitted at regular intervals. Also, since the audio signal is always accompanied in a video call, we do not consider this as an extra overhead.

\vspace{-2pt}
\begin{table}[t]
    \centering
    \setlength{\tabcolsep}{2pt}
    \caption{Quantitative comparison for talking-face video compression on VoxCeleb2 dataset~\cite{voxceleb2_2018_interspeech}. We achieve the best trade-off in terms of quality versus compression ratio. Our method achieves the lowest FID (indicates very high perceptual quality) and a very low/comparable BPP.} 
    
    \vspace{-10pt}
    \begin{tabular}{l|cccc}
    \hline

    \textbf{Method} & \textbf{BPP}$\downarrow$ & \textbf{PSNR}$\uparrow$ & \textbf{SSIM}$\uparrow$& \textbf{FID}$\downarrow$\\
    \hline
    H.264 (CRF=23) (min. compression) & 0.109 & 32.96 & 0.79 & 9.75 \\  
    \hline
    H.264 (CRF=36) & 0.027 & 19.24 & 0.67 & 30.12 \\
    H.266 & 0.0076 & 23.27 & 0.70 & 58.32 \\
    \hline

    fs-vid2vid~\cite{fsvid2vid_nips_2019} & n/a & 20.36 & 0.71 & 85.76\\
    
    os-synth~\cite{nvidia_compression_2021_cvpr} & 0.016 & 24.37 & \textbf{0.80} & 69.13\\
    
    \hline
    \textbf{Ours} & 0.023 & \textbf{24.95} & 0.71 & \textbf{14.10}\\
    \textbf{Ours (Frame-Interpolation)} & \textbf{0.0046} & 23.72 & 0.68 & 14.51 \\

    \hline
    \end{tabular}
    \vspace{-15pt}
    \label{table:compression}
\end{table}

\vspace{5pt}
\noindent
\textbf{Baselines:} 
We benchmark our model's capability using the existing talking-face video compression methods: few-shot vid2vid (fs-vid2vid)~\cite{fsvid2vid_nips_2019}, one-shot free-view synthesis (os-synth)~\cite{nvidia_compression_2021_cvpr} and the standard codecs: (i) H.264 (with CRF of $23$ and $36$) and (ii) H.266 (implemented using vvenc: \url{https://github.com/fraunhoferhhi/vvenc}). Since we train and evaluate on the same dataset (VoxCeleb2~\cite{voxceleb2_2018_interspeech}), we directly take the scores reported in os-synth~\cite{nvidia_compression_2021_cvpr} for comparison.   

\noindent
\textbf{Metrics:}
We compare the compression factor using the standard bits-per-pixels (BPP) metric and measure the reconstruction quality using PSNR, SSIM and FID metrics. 

\begin{figure*}[ht]
  \includegraphics[width=\linewidth]{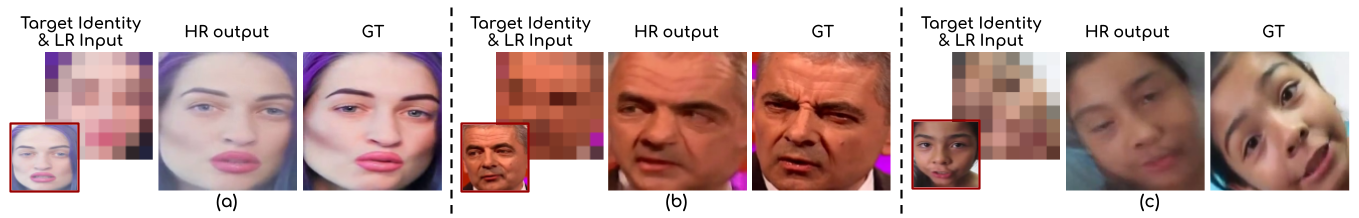}
  \vspace{-20pt}
  \caption{Failure cases: (a) Drastic color changes, (b) Extreme variations of facial expressions and (c) Rapid head motions.}
  \label{fig:limitations}
  \vspace{-10pt}
\end{figure*}

\noindent
\textbf{Results:}
Table~\ref{table:compression} shows the comparison of our approach with the competing methods. We calculate the average BPP across all test videos for H.264 and H.266 codecs. The number of bits required by the current state-of-the-art ``os-synth''~\cite{nvidia_compression_2021_cvpr} to represent a $256\times256$ image is $1056$ (20 keypoints: $(20\times6 + 12)\times8$). Our approach requires $1536$ bits ($8\times8\times3\times8$), with a BPP of $0.023$. The BPP obtained using our method beats H.264 and is comparable to os-synth~\cite{nvidia_compression_2021_cvpr}. In terms of perceptual quality, we can observe that there is a trade-off in compression factor v/s quality for H.264, (i) CRF of $23$: Higher quality, with very less compression and (ii) CRF of $36$: Better compression, but poor reconstruction. While H.266 achieves comparable PSNR and SSIM measures with a BPP of $0.0076$, it still lags behind perceptually as shown by the FID metric. In contrast, our method is able to generate higher quality videos (remarkably low FID), with a better/comparable compression factor. 

\noindent
\textbf{Computation Comparison:}
Our network has $\sim20\%$ fewer parameters ($143$M) while being $2\times$ faster ($50$FPS) than state-of-the-art compression method (os-synth) on an NVIDIA 2080Ti GPU. Previous talking-face generation works focus on obtaining plausible faces, whereas our aim is to develop the first architecture to reconstruct the specific-identity by preserving most of the actual identity details; future works can build upon this to achieve optimal design. Further, model optimization techniques can be applied to the current design to make it suitable for mobile hardware deployment.

\vspace{3pt}
\noindent
\textbf{Frame-Interpolation Network:}
To further reduce the bandwidth consumption, we develop a baseline frame-interpolation network with an aim to upsample $5$FPS videos to $25$FPS. While such networks have been studied in literature~\cite{super-slowmo}, most of them restrict the amount of upsampling, mainly due to the computational resources involved in training the bulky networks (often consisting of $3$D CNNs). In contrast, we take the advantage of LR videos and design a model to increase the temporal resolution of frames ($5$FPS to $25$FPS) for the first time. Our encoder-decoder based model processes $5$FPS LR frames ($5,3,8,8$) and upsamples it $5\times$ to generate $25$FPS LR frames as output ($25,3,8,8$). We refer the reader to our supplementary for more details. The lower resolution of both the input and output allows us to train a very light $3$D CNN model with only $0.2M$ parameters. This strategy effectively permits us to transmit $5$FPS LR videos. On receiving these frames, frame-interpolation network initially upsamples them to $25$FPS LR videos, which can subsequently be ingested by our spatial talking-face video upsampling network to render the final HR video. As shown in Table~\ref{table:compression}, our frame-interpolation network achieves a further reduction in BPP, since only $1$ in $5$ LR frames needs to be transferred. We thus obtain $\sim6\times$ and $\sim25\times$ reduction in bandwidth compared to os-synth~\cite{nvidia_compression_2021_cvpr} and H.264 codec (with CRF $23$) respectively, without a significant compromise in the generated quality.

\vspace{-5pt}
\subsection{Audio-driven Talking-Face Generation (A2TF) and Face Re-enactment (FR)}
\label{section:A2TH_comparison}

\noindent
\textbf{Baselines:}
For A2TF, we compare with Wav2Lip~\cite{wav2lip} and MakeItTalk \cite{Yang:2020:MakeItTalk}. Talking-face videos are generated using the audio segment and the first frame of the original HD video as inputs (same strategy as our model). For FR, the original models take the actual HD frames as the driving video input, however, we do not have access to the actual HD video in our model. Thus, we upsample the LR input using the existing video SR method, TecoGAN~\cite{tecogan} (trained on VoxCeleb2 dataset) and use it as our input driving video, along with the first frame of the original HD video as target identity input. 

\noindent
\textbf{Metrics:}
Along with the standard metrics, we specifically evaluate the ability of models to match the original identity using head pose estimation metrics~\cite{LwPosr_wacv_2022, 6DHeadPose_arxiv_2022}. We convert the rotation matrix\footnote{\url{https://learnopencv.com/head-pose-estimation-using-opencv-and-dlib/}} to Euler angles and report Mean Absolute Error (MAE) of these angles. 

\noindent
\textbf{Results:}
Table~\ref{table:a2tf_fr_comparison} shows the performance of different A2TF and FR methods in comparison to our approach. A2TF models generate accurate lip-sync (LSE-D), but they fail to match the pose of the original identity (as noted by Yaw, Pitch, Roll and MAE metrics). FR methods also face similar issues and do not match the identity, as they do not have the actual HD video as driving frames. Our model is clearly the most applicable for this task, is consistent across all metrics and produces noticeably better results. Most importantly, our model preserves the identity to the maximum extent. 

\vspace{-9pt}
\begin{table}[ht]
    \centering
    \footnotesize
    \setlength{\tabcolsep}{1.2pt}
    \caption{Quantitative comparison with A2TF~\cite{wav2lip, Yang:2020:MakeItTalk} and FR~\cite{fomm_nips_2019, PC_AVS_2021_cvpr} methods on VoxCeleb2 test set.}

    \vspace{-10pt}
    \begin{tabular}{l|ccccccccccc}
    \hline
    
    \textbf{Method} & 
    \textbf{PSNR}$\uparrow$ & \textbf{SSIM}$\uparrow$ & \textbf{FID}$\downarrow$ & \textbf{LMD}$\downarrow$ & \textbf{LSE-D}$\downarrow$ & \textbf{Yaw}$\downarrow$ & \textbf{Pitch}$\downarrow$ & \textbf{Roll}$\downarrow$ & \textbf{MAE}$\downarrow$\\
     
    \hline
    
    Wav2Lip~\cite{wav2lip} & 14.18 & 0.32 & \textbf{8.15} & 4.320 & 9.19 & 24.68 & 38.31 & 28.94 & 30.64 \\
    
    MakeItTalk~\cite{Yang:2020:MakeItTalk} & 18.88 & 0.49 & 31.19 & 2.012 & 11.91 & 26.29 & 40.13 & 31.42 & 32.61 \\
    
    \hline
    
    FOMM~\cite{fomm_nips_2019} & 20.14 & 0.56 & 21.18 & 0.864 & 14.03 & 19.14 & 30.57 & 22.76 & 24.35 \\
    
    PC-AVS~\cite{PC_AVS_2021_cvpr} & 15.68 & 0.37 & 33.38 & 1.063 & \textbf{8.42} & 22.27 & 31.89 & 25.80 & 26.65 \\
        
    \hline
    
    \textbf{Ours} & \textbf{24.95} & \textbf{0.71} & 14.10 & \textbf{0.196} & 13.91 & \textbf{13.55} & \textbf{21.01} & \textbf{15.48} & \textbf{16.68}\\

    \hline
    
    \end{tabular}
    \vspace{-10pt}
    \label{table:a2tf_fr_comparison}
\end{table}

\vspace{-5pt}
\section{Limitations and Future Directions}

Although our method generates realistic results for a wide variety of inputs, there are certain situations as shown in Figure~\ref{fig:limitations} where our model results in sub-optimal generations. For example, if the color contrast changes drastically as the video progresses, the model fails to capture these details (Figure~\ref{fig:limitations} (a)). Significant variations in facial expressions is another case where our model struggles to replicate the details (Figure~\ref{fig:limitations} (b)). Explicitly handling the expressions is an interesting direction that can be investigated in the future, which we currently do not handle in our work. Finally, in the case of sudden/rapid changes in view, camera angle, or head movements, our model attempts to generate smooth transitions (Figure~\ref{fig:limitations} (c)). However, we found our method to be stable over a large variety of inputs and anticipate that our idea of utilizing extremely LR frames will be a basis for other domains and applications.

\vspace{-5pt}
\section{Conclusion}
\label{section:conclusion}

In this work, we present a novel framework for extreme-scale talking-face video super-resolution and compression. We show that by considering appropriate priors (audio signal and a single target identity image), we are able to generate realistic, high-quality talking-face videos from (very) low-resolution frames. Our method handles various inputs, including but not limited to people of different ages, gender and ethnicity. Most importantly, our framework is the first of its kind to produce photo-realistic lip-synced talking-face videos while also matching the actual identity. By dramatically reducing the bandwidth requirements, our approach can be utilized as a tool for a seamless video-conferencing experience. We believe our core idea of exploiting very low-resolution videos along with adequate priors will be an important step towards the future of super-resolution and low-bandwidth video-conferencing.

\bibliographystyle{ACM-Reference-Format}
\balance
\bibliography{sample-base}


\begin{thebibliography}{58}


\ifx \showCODEN    \undefined \def \showCODEN     #1{\unskip}     \fi
\ifx \showDOI      \undefined \def \showDOI       #1{#1}\fi
\ifx \showISBNx    \undefined \def \showISBNx     #1{\unskip}     \fi
\ifx \showISBNxiii \undefined \def \showISBNxiii  #1{\unskip}     \fi
\ifx \showISSN     \undefined \def \showISSN      #1{\unskip}     \fi
\ifx \showLCCN     \undefined \def \showLCCN      #1{\unskip}     \fi
\ifx \shownote     \undefined \def \shownote      #1{#1}          \fi
\ifx \showarticletitle \undefined \def \showarticletitle #1{#1}   \fi
\ifx \showURL      \undefined \def \showURL       {\relax}        \fi
\providecommand\bibfield[2]{#2}
\providecommand\bibinfo[2]{#2}
\providecommand\natexlab[1]{#1}
\providecommand\showeprint[2][]{arXiv:#2}

\bibitem[Afouras et~al\mbox{.}(2018)]%
        {conversation}
\bibfield{author}{\bibinfo{person}{T. Afouras}, \bibinfo{person}{J.~S. Chung},
  {and} \bibinfo{person}{A. Zisserman}.} \bibinfo{year}{2018}\natexlab{}.
\newblock \showarticletitle{The Conversation: Deep Audio-Visual Speech
  Enhancement}. In \bibinfo{booktitle}{\emph{INTERSPEECH}}.
\newblock


\bibitem[Afouras et~al\mbox{.}(2020)]%
        {avobjects}
\bibfield{author}{\bibinfo{person}{Triantafyllos Afouras},
  \bibinfo{person}{Andrew Owens}, \bibinfo{person}{Joon~Son Chung}, {and}
  \bibinfo{person}{Andrew Zisserman}.} \bibinfo{year}{2020}\natexlab{}.
\newblock \showarticletitle{Self-Supervised Learning of Audio-Visual Objects
  from Video}. In \bibinfo{booktitle}{\emph{European Conference on Computer
  Vision}}.
\newblock


\bibitem[Agustsson et~al\mbox{.}(2020)]%
        {Agustsson_2020_CVPR}
\bibfield{author}{\bibinfo{person}{Eirikur Agustsson}, \bibinfo{person}{David
  Minnen}, \bibinfo{person}{Nick Johnston}, \bibinfo{person}{Johannes Balle},
  \bibinfo{person}{Sung~Jin Hwang}, {and} \bibinfo{person}{George Toderici}.}
  \bibinfo{year}{2020}\natexlab{}.
\newblock \showarticletitle{Scale-Space Flow for End-to-End Optimized Video
  Compression}. In \bibinfo{booktitle}{\emph{Proceedings of the IEEE/CVF
  Conference on Computer Vision and Pattern Recognition (CVPR)}}.
\newblock


\bibitem[Bulat and Tzimiropoulos(2017)]%
        {fan_landmark_iccv_2017}
\bibfield{author}{\bibinfo{person}{Adrian Bulat} {and}
  \bibinfo{person}{Georgios Tzimiropoulos}.} \bibinfo{year}{2017}\natexlab{}.
\newblock \showarticletitle{How far are we from solving the 2D \& 3D Face
  Alignment problem? (and a dataset of 230,000 3D facial landmarks)}. In
  \bibinfo{booktitle}{\emph{International Conference on Computer Vision}}.
\newblock


\bibitem[Bulat et~al\mbox{.}(2018)]%
        {face_sr_somework}
\bibfield{author}{\bibinfo{person}{Adrian Bulat}, \bibinfo{person}{Jing Yang},
  {and} \bibinfo{person}{Georgios Tzimiropoulos}.}
  \bibinfo{year}{2018}\natexlab{}.
\newblock \showarticletitle{To learn image super-resolution, use a GAN to learn
  how to do image degradation first}. In \bibinfo{booktitle}{\emph{Proceedings
  of the European Conference on Computer Vision (ECCV)}}.
  \bibinfo{pages}{185--200}.
\newblock


\bibitem[Chen et~al\mbox{.}(2020)]%
        {SPARNet}
\bibfield{author}{\bibinfo{person}{Chaofeng Chen}, \bibinfo{person}{Dihong
  Gong}, \bibinfo{person}{Hao Wang}, \bibinfo{person}{Zhifeng Li}, {and}
  \bibinfo{person}{Kwan-Yee~K. Wong}.} \bibinfo{year}{2020}\natexlab{}.
\newblock \showarticletitle{Learning Spatial Attention for Face
  Super-Resolution}.
\newblock \bibinfo{journal}{\emph{IEEE Transactions on Image Processing
  (TIP)}}.
\newblock


\bibitem[Chu et~al\mbox{.}(2020)]%
        {tecogan}
\bibfield{author}{\bibinfo{person}{Mengyu Chu}, \bibinfo{person}{You Xie},
  \bibinfo{person}{Jonas Mayer}, \bibinfo{person}{Laura Leal-Taix\'{e}}, {and}
  \bibinfo{person}{Nils Thuerey}.} \bibinfo{year}{2020}\natexlab{}.
\newblock \showarticletitle{Learning Temporal Coherence via Self-Supervision
  for GAN-Based Video Generation}.
\newblock \bibinfo{journal}{\emph{ACM Trans. Graph.}}  \bibinfo{volume}{39}
  (\bibinfo{date}{July} \bibinfo{year}{2020}), \bibinfo{numpages}{13}~pages.
\newblock
\urldef\tempurl%
\url{https://doi.org/10.1145/3386569.3392457}
\showDOI{\tempurl}


\bibitem[Chung et~al\mbox{.}(2018)]%
        {voxceleb2_2018_interspeech}
\bibfield{author}{\bibinfo{person}{J.~S. Chung}, \bibinfo{person}{A. Nagrani},
  {and} \bibinfo{person}{A. Zisserman}.} \bibinfo{year}{2018}\natexlab{}.
\newblock \showarticletitle{VoxCeleb2: Deep Speaker Recognition}. In
  \bibinfo{booktitle}{\emph{INTERSPEECH}}.
\newblock


\bibitem[Chung and Zisserman(2016)]%
        {syncnet_2016}
\bibfield{author}{\bibinfo{person}{J.~S. Chung} {and} \bibinfo{person}{A.
  Zisserman}.} \bibinfo{year}{2016}\natexlab{}.
\newblock \showarticletitle{Out of time: automated lip sync in the wild}. In
  \bibinfo{booktitle}{\emph{Workshop on Multi-view Lip-reading, ACCV}}.
\newblock


\bibitem[{Chung} et~al\mbox{.}(2019)]%
        {perfect_match}
\bibfield{author}{\bibinfo{person}{S. {Chung}}, \bibinfo{person}{J.~S.
  {Chung}}, {and} \bibinfo{person}{H. {Kang}}.}
  \bibinfo{year}{2019}\natexlab{}.
\newblock \showarticletitle{Perfect Match: Improved Cross-modal Embeddings for
  Audio-visual Synchronisation}. In \bibinfo{booktitle}{\emph{ICASSP 2019 -
  2019 IEEE International Conference on Acoustics, Speech and Signal Processing
  (ICASSP)}}. \bibinfo{pages}{3965--3969}.
\newblock
\urldef\tempurl%
\url{https://doi.org/10.1109/ICASSP.2019.8682524}
\showDOI{\tempurl}


\bibitem[de~Barros~Reis et~al\mbox{.}(2020)]%
        {talking_face_video_sig}
\bibfield{author}{\bibinfo{person}{Filipe~Antonio de Barros~Reis},
  \bibinfo{person}{Paula Dornhofer Paro~Costa}, {and}
  \bibinfo{person}{Jos\'{e}~Mario de Martino}.}
  \bibinfo{year}{2020}\natexlab{}.
\newblock \showarticletitle{Deeply Emotional Talking Head: A Generative
  Adversarial Network Approach to Expressive Speech Synthesis with Emotion
  Control}. In \bibinfo{booktitle}{\emph{ACM SIGGRAPH 2020 Posters}} (Virtual
  Event, USA) \emph{(\bibinfo{series}{SIGGRAPH '20})}.
  \bibinfo{publisher}{Association for Computing Machinery},
  \bibinfo{address}{New York, NY, USA}, Article \bibinfo{articleno}{33},
  \bibinfo{numpages}{2}~pages.
\newblock
\showISBNx{9781450379731}
\urldef\tempurl%
\url{https://doi.org/10.1145/3388770.3407417}
\showDOI{\tempurl}


\bibitem[Dhingra(2022)]%
        {LwPosr_wacv_2022}
\bibfield{author}{\bibinfo{person}{Naina Dhingra}.}
  \bibinfo{year}{2022}\natexlab{}.
\newblock \showarticletitle{LwPosr: Lightweight Efficient Fine Grained Head
  Pose Estimation}. In \bibinfo{booktitle}{\emph{Proceedings of the IEEE/CVF
  Winter Conference on Applications of Computer Vision}}.
  \bibinfo{pages}{1495--1505}.
\newblock


\bibitem[Dong et~al\mbox{.}(2016)]%
        {dong_first_sr}
\bibfield{author}{\bibinfo{person}{Chao Dong}, \bibinfo{person}{Chen~Change
  Loy}, \bibinfo{person}{Kaiming He}, {and} \bibinfo{person}{Xiaoou Tang}.}
  \bibinfo{year}{2016}\natexlab{}.
\newblock \showarticletitle{Image Super-Resolution Using Deep Convolutional
  Networks}.
\newblock \bibinfo{journal}{\emph{IEEE Transactions on Pattern Analysis and
  Machine Intelligence}} \bibinfo{volume}{38}, \bibinfo{number}{2}
  (\bibinfo{year}{2016}), \bibinfo{pages}{295--307}.
\newblock
\urldef\tempurl%
\url{https://doi.org/10.1109/TPAMI.2015.2439281}
\showDOI{\tempurl}


\bibitem[Dumas et~al\mbox{.}(2018)]%
        {CompressionAE_icassp_2018}
\bibfield{author}{\bibinfo{person}{Thierry Dumas}, \bibinfo{person}{Aline
  Roumy}, {and} \bibinfo{person}{Christine Guillemot}.}
  \bibinfo{year}{2018}\natexlab{}.
\newblock \showarticletitle{Autoencoder Based Image Compression: Can the
  Learning be Quantization Independent?}
\newblock \bibinfo{journal}{\emph{IEEE International Conference on Acoustics,
  Speech and Signal Processing (ICASSP)}} (\bibinfo{year}{2018}),
  \bibinfo{pages}{1188--1192}.
\newblock


\bibitem[Efros et~al\mbox{.}(2003)]%
        {efros}
\bibfield{author}{\bibinfo{person}{Efros}, \bibinfo{person}{Berg},
  \bibinfo{person}{Mori}, {and} \bibinfo{person}{Malik}.}
  \bibinfo{year}{2003}\natexlab{}.
\newblock \showarticletitle{Recognizing action at a distance}. In
  \bibinfo{booktitle}{\emph{Proceedings Ninth IEEE International Conference on
  Computer Vision}}. \bibinfo{pages}{726--733 vol.2}.
\newblock
\urldef\tempurl%
\url{https://doi.org/10.1109/ICCV.2003.1238420}
\showDOI{\tempurl}


\bibitem[Ephrat et~al\mbox{.}(2018)]%
        {avspeech_2018_tog}
\bibfield{author}{\bibinfo{person}{Ariel Ephrat}, \bibinfo{person}{Inbar
  Mosseri}, \bibinfo{person}{Oran Lang}, \bibinfo{person}{Tali Dekel},
  \bibinfo{person}{Kevin Wilson}, \bibinfo{person}{Avinatan Hassidim},
  \bibinfo{person}{William~T. Freeman}, {and} \bibinfo{person}{Michael
  Rubinstein}.} \bibinfo{year}{2018}\natexlab{}.
\newblock \showarticletitle{Looking to Listen at the Cocktail Party: A
  Speaker-Independent Audio-Visual Model for Speech Separation}.
\newblock \bibinfo{journal}{\emph{ACM Trans. Graph.}} \bibinfo{volume}{37},
  \bibinfo{number}{4}, Article \bibinfo{articleno}{112} (\bibinfo{year}{2018}),
  \bibinfo{numpages}{11}~pages.
\newblock
\showISSN{0730-0301}
\urldef\tempurl%
\url{https://doi.org/10.1145/3197517.3201357}
\showDOI{\tempurl}


\bibitem[Fried et~al\mbox{.}(2019)]%
        {text-based-editing}
\bibfield{author}{\bibinfo{person}{Ohad Fried}, \bibinfo{person}{Ayush Tewari},
  \bibinfo{person}{Michael Zollh\"{o}fer}, \bibinfo{person}{Adam Finkelstein},
  \bibinfo{person}{Eli Shechtman}, \bibinfo{person}{Dan~B Goldman},
  \bibinfo{person}{Kyle Genova}, \bibinfo{person}{Zeyu Jin},
  \bibinfo{person}{Christian Theobalt}, {and} \bibinfo{person}{Maneesh
  Agrawala}.} \bibinfo{year}{2019}\natexlab{}.
\newblock \showarticletitle{Text-Based Editing of Talking-Head Video}.
\newblock \bibinfo{journal}{\emph{ACM Trans. Graph.}}  \bibinfo{volume}{38},
  Article \bibinfo{articleno}{68} (\bibinfo{year}{2019}),
  \bibinfo{numpages}{14}~pages.
\newblock
\urldef\tempurl%
\url{https://doi.org/10.1145/3306346.3323028}
\showDOI{\tempurl}


\bibitem[Haris et~al\mbox{.}(2018)]%
        {dbpn}
\bibfield{author}{\bibinfo{person}{Muhammad Haris}, \bibinfo{person}{Greg
  Shakhnarovich}, {and} \bibinfo{person}{Norimichi Ukita}.}
  \bibinfo{year}{2018}\natexlab{}.
\newblock \showarticletitle{Deep Back-Projection Networks for
  Super-Resolution}. In \bibinfo{booktitle}{\emph{2018 IEEE/CVF Conference on
  Computer Vision and Pattern Recognition}}. \bibinfo{pages}{1664--1673}.
\newblock
\urldef\tempurl%
\url{https://doi.org/10.1109/CVPR.2018.00179}
\showDOI{\tempurl}


\bibitem[Haris et~al\mbox{.}(2019)]%
        {RBPN2019}
\bibfield{author}{\bibinfo{person}{Muhammad Haris}, \bibinfo{person}{Greg
  Shakhnarovich}, {and} \bibinfo{person}{Norimichi Ukita}.}
  \bibinfo{year}{2019}\natexlab{}.
\newblock \showarticletitle{Recurrent Back-Projection Network for Video
  Super-Resolution}. In \bibinfo{booktitle}{\emph{IEEE Conference on Computer
  Vision and Pattern Recognition (CVPR)}}.
\newblock


\bibitem[H{\'{e}}as et~al\mbox{.}(2015)]%
        {video_sr_1}
\bibfield{author}{\bibinfo{person}{Patrick H{\'{e}}as},
  \bibinfo{person}{Ang{\'{e}}lique Dr{\'{e}}meau}, {and}
  \bibinfo{person}{C{\'{e}}dric Herzet}.} \bibinfo{year}{2015}\natexlab{}.
\newblock \showarticletitle{Estimation of Super-Resolved Video Dynamics}.
\newblock \bibinfo{journal}{\emph{CoRR}}  \bibinfo{volume}{abs/1506.00473}
  (\bibinfo{year}{2015}).
\newblock
\showeprint[arxiv]{1506.00473}


\bibitem[Hegde et~al\mbox{.}(2021)]%
        {Hegde_2021_WACV}
\bibfield{author}{\bibinfo{person}{Sindhu~B. Hegde}, \bibinfo{person}{K.R.
  Prajwal}, \bibinfo{person}{Rudrabha Mukhopadhyay}, \bibinfo{person}{Vinay~P.
  Namboodiri}, {and} \bibinfo{person}{C.V. Jawahar}.}
  \bibinfo{year}{2021}\natexlab{}.
\newblock \showarticletitle{Visual Speech Enhancement Without a Real Visual
  Stream}. In \bibinfo{booktitle}{\emph{Proceedings of the IEEE/CVF Winter
  Conference on Applications of Computer Vision (WACV)}}.
  \bibinfo{pages}{1926--1935}.
\newblock


\bibitem[Hempel et~al\mbox{.}(2022)]%
        {6DHeadPose_arxiv_2022}
\bibfield{author}{\bibinfo{person}{Thorsten Hempel}, \bibinfo{person}{Ahmed~A
  Abdelrahman}, {and} \bibinfo{person}{Ayoub Al-Hamadi}.}
  \bibinfo{year}{2022}\natexlab{}.
\newblock \showarticletitle{6D Rotation Representation For Unconstrained Head
  Pose Estimation}.
\newblock \bibinfo{journal}{\emph{arXiv preprint arXiv:2202.12555}}
  (\bibinfo{year}{2022}).
\newblock


\bibitem[Heusel et~al\mbox{.}(2018)]%
        {heusel2018gans}
\bibfield{author}{\bibinfo{person}{Martin Heusel}, \bibinfo{person}{Hubert
  Ramsauer}, \bibinfo{person}{Thomas Unterthiner}, \bibinfo{person}{Bernhard
  Nessler}, {and} \bibinfo{person}{Sepp Hochreiter}.}
  \bibinfo{year}{2018}\natexlab{}.
\newblock \bibinfo{title}{GANs Trained by a Two Time-Scale Update Rule Converge
  to a Local Nash Equilibrium}.
\newblock
\newblock
\showeprint[arxiv]{1706.08500}~[cs.LG]


\bibitem[Ho et~al\mbox{.}(2019)]%
        {CompressionFlow_nips_2019}
\bibfield{author}{\bibinfo{person}{Jonathan Ho}, \bibinfo{person}{Evan Lohn},
  {and} \bibinfo{person}{P. Abbeel}.} \bibinfo{year}{2019}\natexlab{}.
\newblock \showarticletitle{Compression with Flows via Local Bits-Back Coding}.
  In \bibinfo{booktitle}{\emph{NeurIPS}}.
\newblock


\bibitem[Hu and Ramanan(2017)]%
        {Hu_2017_CVPR}
\bibfield{author}{\bibinfo{person}{Peiyun Hu} {and} \bibinfo{person}{Deva
  Ramanan}.} \bibinfo{year}{2017}\natexlab{}.
\newblock \showarticletitle{Finding Tiny Faces}. In
  \bibinfo{booktitle}{\emph{The IEEE Conference on Computer Vision and Pattern
  Recognition (CVPR)}}.
\newblock


\bibitem[Jamaludin et~al\mbox{.}(2019)]%
        {YST_ijcv_2019}
\bibfield{author}{\bibinfo{person}{Amir Jamaludin}, \bibinfo{person}{Joon~Son
  Chung}, {and} \bibinfo{person}{Andrew Zisserman}.}
  \bibinfo{year}{2019}\natexlab{}.
\newblock \showarticletitle{You said that? : Synthesising talking faces from
  audio}.
\newblock \bibinfo{journal}{\emph{International Journal of Computer Vision}}
  (\bibinfo{year}{2019}).
\newblock


\bibitem[Jiang et~al\mbox{.}(2018)]%
        {super-slowmo}
\bibfield{author}{\bibinfo{person}{Huaizu Jiang}, \bibinfo{person}{Deqing Sun},
  \bibinfo{person}{V. Jampani}, \bibinfo{person}{Ming-Hsuan Yang},
  \bibinfo{person}{Erik~G. Learned-Miller}, {and} \bibinfo{person}{Jan Kautz}.}
  \bibinfo{year}{2018}\natexlab{}.
\newblock \showarticletitle{Super SloMo: High Quality Estimation of Multiple
  Intermediate Frames for Video Interpolation}.
\newblock \bibinfo{journal}{\emph{2018 IEEE/CVF Conference on Computer Vision
  and Pattern Recognition}} (\bibinfo{year}{2018}),
  \bibinfo{pages}{9000--9008}.
\newblock


\bibitem[K~R et~al\mbox{.}(2019)]%
        {lipgan_2019}
\bibfield{author}{\bibinfo{person}{Prajwal K~R}, \bibinfo{person}{Rudrabha
  Mukhopadhyay}, \bibinfo{person}{Jerin Philip}, \bibinfo{person}{Abhishek
  Jha}, \bibinfo{person}{Vinay Namboodiri}, {and} \bibinfo{person}{C~V
  Jawahar}.} \bibinfo{year}{2019}\natexlab{}.
\newblock \showarticletitle{Towards Automatic Face-to-Face Translation}. In
  \bibinfo{booktitle}{\emph{Proceedings of the 27th ACM International
  Conference on Multimedia}} (Nice, France) \emph{(\bibinfo{series}{MM '19})}.
  \bibinfo{publisher}{ACM}, \bibinfo{numpages}{9}~pages.
\newblock
\urldef\tempurl%
\url{https://doi.org/10.1145/3343031.3351066}
\showDOI{\tempurl}


\bibitem[Khani et~al\mbox{.}(2021)]%
        {SRVC_2021_ICCV}
\bibfield{author}{\bibinfo{person}{Mehrdad Khani},
  \bibinfo{person}{Vibhaalakshmi Sivaraman}, {and} \bibinfo{person}{Mohammad
  Alizadeh}.} \bibinfo{year}{2021}\natexlab{}.
\newblock \showarticletitle{Efficient Video Compression via Content-Adaptive
  Super-Resolution}. In \bibinfo{booktitle}{\emph{Proceedings of the IEEE/CVF
  International Conference on Computer Vision (ICCV)}}.
  \bibinfo{pages}{4521--4530}.
\newblock


\bibitem[Kim et~al\mbox{.}(2018b)]%
        {FaceVoiceAssociations_accv_2018}
\bibfield{author}{\bibinfo{person}{Changil Kim},
  \bibinfo{person}{Hijung~Valentina Shin}, \bibinfo{person}{Tae-Hyun Oh},
  \bibinfo{person}{Alexandre Kaspar}, \bibinfo{person}{Mohamed Elgharib}, {and}
  \bibinfo{person}{Wojciech Matusik}.} \bibinfo{year}{2018}\natexlab{b}.
\newblock \showarticletitle{On Learning Associations of Faces and Voices}. In
  \bibinfo{booktitle}{\emph{Proceedings of Asian Conference on Computer Vision
  (ACCV)}}.
\newblock


\bibitem[Kim et~al\mbox{.}(2019)]%
        {progressive-face-sr}
\bibfield{author}{\bibinfo{person}{Deokyun Kim}, \bibinfo{person}{Minseon Kim},
  \bibinfo{person}{Gihyun Kwon}, {and} \bibinfo{person}{Dae-Shik Kim}.}
  \bibinfo{year}{2019}\natexlab{}.
\newblock \showarticletitle{Progressive Face Super-Resolution via Attention to
  Facial Landmark}. In \bibinfo{booktitle}{\emph{Proceedings of the 30th
  British Machine Vision Conference (BMVC)}}.
\newblock


\bibitem[Kim et~al\mbox{.}(2018a)]%
        {deep_video_portraits}
\bibfield{author}{\bibinfo{person}{H. Kim}, \bibinfo{person}{Pablo Garrido},
  \bibinfo{person}{Ayush Tewari}, \bibinfo{person}{Weipeng Xu},
  \bibinfo{person}{Justus Thies}, \bibinfo{person}{Matthias Nie{\ss}ner},
  \bibinfo{person}{P. P{\'e}rez}, \bibinfo{person}{Christian Richardt},
  \bibinfo{person}{M. Zollh{\"o}fer}, {and} \bibinfo{person}{C. Theobalt}.}
  \bibinfo{year}{2018}\natexlab{a}.
\newblock \showarticletitle{Deep video portraits}.
\newblock \bibinfo{journal}{\emph{ACM Transactions on Graphics (TOG)}}
  \bibinfo{volume}{37} (\bibinfo{year}{2018}), \bibinfo{pages}{1 -- 14}.
\newblock


\bibitem[Kim et~al\mbox{.}(2015)]%
        {vdsr}
\bibfield{author}{\bibinfo{person}{Jiwon Kim}, \bibinfo{person}{Jung~Kwon Lee},
  {and} \bibinfo{person}{Kyoung~Mu Lee}.} \bibinfo{year}{2015}\natexlab{}.
\newblock \showarticletitle{Accurate Image Super-Resolution Using Very Deep
  Convolutional Networks}.
\newblock \bibinfo{journal}{\emph{CoRR}}  \bibinfo{volume}{abs/1511.04587}
  (\bibinfo{year}{2015}).
\newblock
\showeprint[arxiv]{1511.04587}
\urldef\tempurl%
\url{http://arxiv.org/abs/1511.04587}
\showURL{%
\tempurl}


\bibitem[Kim et~al\mbox{.}(2021)]%
        {KIM202111}
\bibfield{author}{\bibinfo{person}{Jonghyun Kim}, \bibinfo{person}{Gen Li},
  \bibinfo{person}{Inyong Yun}, \bibinfo{person}{Cheolkon Jung}, {and}
  \bibinfo{person}{Joongkyu Kim}.} \bibinfo{year}{2021}\natexlab{}.
\newblock \showarticletitle{Edge and identity preserving network for face
  super-resolution}.
\newblock \bibinfo{journal}{\emph{Neurocomputing}}  \bibinfo{volume}{446}
  (\bibinfo{year}{2021}), \bibinfo{pages}{11--22}.
\newblock
\urldef\tempurl%
\url{https://doi.org/10.1016/j.neucom.2021.03.048}
\showDOI{\tempurl}


\bibitem[Lim et~al\mbox{.}(2017)]%
        {edsr}
\bibfield{author}{\bibinfo{person}{Bee Lim}, \bibinfo{person}{Sanghyun Son},
  \bibinfo{person}{Heewon Kim}, \bibinfo{person}{Seungjun Nah}, {and}
  \bibinfo{person}{Kyoung~Mu Lee}.} \bibinfo{year}{2017}\natexlab{}.
\newblock \showarticletitle{Enhanced Deep Residual Networks for Single Image
  Super-Resolution}. In \bibinfo{booktitle}{\emph{2017 IEEE Conference on
  Computer Vision and Pattern Recognition Workshops (CVPRW)}}.
  \bibinfo{pages}{1132--1140}.
\newblock
\urldef\tempurl%
\url{https://doi.org/10.1109/CVPRW.2017.151}
\showDOI{\tempurl}


\bibitem[Ma et~al\mbox{.}(2020)]%
        {dicgan}
\bibfield{author}{\bibinfo{person}{Cheng Ma}, \bibinfo{person}{Zhenyu Jiang},
  \bibinfo{person}{Yongming Rao}, \bibinfo{person}{Jiwen Lu}, {and}
  \bibinfo{person}{J. Zhou}.} \bibinfo{year}{2020}\natexlab{}.
\newblock \showarticletitle{Deep Face Super-Resolution With Iterative
  Collaboration Between Attentive Recovery and Landmark Estimation}.
\newblock \bibinfo{journal}{\emph{2020 IEEE/CVF Conference on Computer Vision
  and Pattern Recognition (CVPR)}} (\bibinfo{year}{2020}),
  \bibinfo{pages}{5568--5577}.
\newblock


\bibitem[Menon et~al\mbox{.}(2020)]%
        {PULSE_CVPR_2020}
\bibfield{author}{\bibinfo{person}{Sachit Menon}, \bibinfo{person}{Alex
  Damian}, \bibinfo{person}{McCourt Hu}, \bibinfo{person}{Nikhil Ravi}, {and}
  \bibinfo{person}{Cynthia Rudin}.} \bibinfo{year}{2020}\natexlab{}.
\newblock \showarticletitle{PULSE: Self-Supervised Photo Upsampling via Latent
  Space Exploration of Generative Models}. In \bibinfo{booktitle}{\emph{The
  IEEE Conference on Computer Vision and Pattern Recognition (CVPR)}}.
\newblock


\bibitem[Nagrani et~al\mbox{.}(2018)]%
        {SeeingVoices_cvpr_2018}
\bibfield{author}{\bibinfo{person}{Arsha Nagrani}, \bibinfo{person}{Samuel
  Albanie}, {and} \bibinfo{person}{Andrew Zisserman}.}
  \bibinfo{year}{2018}\natexlab{}.
\newblock \showarticletitle{Seeing Voices and Hearing Faces: Cross-Modal
  Biometric Matching}.
\newblock \bibinfo{journal}{\emph{IEEE/CVF Conference on Computer Vision and
  Pattern Recognition (CVPR)}} (\bibinfo{year}{2018}),
  \bibinfo{pages}{8427--8436}.
\newblock


\bibitem[Narayanan and Kiong(2013)]%
        {VideoConfMedical_2013_atnac}
\bibfield{author}{\bibinfo{person}{Arun~Shankar Narayanan} {and}
  \bibinfo{person}{Tan~Kok Kiong}.} \bibinfo{year}{2013}\natexlab{}.
\newblock \showarticletitle{Video conferencing solution for medical
  applications in low-bandwidth networks}. In \bibinfo{booktitle}{\emph{2013
  Australasian Telecommunication Networks and Applications Conference
  (ATNAC)}}. \bibinfo{pages}{195--200}.
\newblock
\urldef\tempurl%
\url{https://doi.org/10.1109/ATNAC.2013.6705380}
\showDOI{\tempurl}


\bibitem[Oh et~al\mbox{.}(2019)]%
        {speech2face}
\bibfield{author}{\bibinfo{person}{Tae-Hyun Oh}, \bibinfo{person}{Tali Dekel},
  \bibinfo{person}{Changil Kim}, \bibinfo{person}{Inbar Mosseri},
  \bibinfo{person}{William~T. Freeman}, \bibinfo{person}{Michael Rubinstein},
  {and} \bibinfo{person}{Wojciech Matusik}.} \bibinfo{year}{2019}\natexlab{}.
\newblock \showarticletitle{Speech2Face: Learning the Face Behind a Voice}. In
  \bibinfo{booktitle}{\emph{IEEE/CVF Conference on Computer Vision and Pattern
  Recognition (CVPR)}}. \bibinfo{pages}{7531--7540}.
\newblock
\urldef\tempurl%
\url{https://doi.org/10.1109/CVPR.2019.00772}
\showDOI{\tempurl}


\bibitem[Oquab et~al\mbox{.}(2021)]%
        {VideoChatCompression_2021_cvprw}
\bibfield{author}{\bibinfo{person}{Maxime Oquab}, \bibinfo{person}{Pierre
  Stock}, \bibinfo{person}{Oran Gafni}, \bibinfo{person}{Daniel Haziza},
  \bibinfo{person}{Tao Xu}, \bibinfo{person}{Peizhao Zhang},
  \bibinfo{person}{Onur Çelebi}, \bibinfo{person}{Yana Hasson},
  \bibinfo{person}{Patrick Labatut}, \bibinfo{person}{Bobo Bose-Kolanu},
  \bibinfo{person}{Thibault Peyronel}, {and} \bibinfo{person}{Camille
  Couprie}.} \bibinfo{year}{2021}\natexlab{}.
\newblock \showarticletitle{Low Bandwidth Video-Chat Compression using Deep
  Generative Models}.
\newblock \bibinfo{journal}{\emph{2021 IEEE/CVF Conference on Computer Vision
  and Pattern Recognition Workshops (CVPRW)}} (\bibinfo{year}{2021}),
  \bibinfo{pages}{2388--2397}.
\newblock


\bibitem[Prajwal et~al\mbox{.}(2020)]%
        {wav2lip}
\bibfield{author}{\bibinfo{person}{K~R Prajwal}, \bibinfo{person}{Rudrabha
  Mukhopadhyay}, \bibinfo{person}{Vinay~P. Namboodiri}, {and}
  \bibinfo{person}{C.V. Jawahar}.} \bibinfo{year}{2020}\natexlab{}.
\newblock \showarticletitle{A Lip Sync Expert Is All You Need for Speech to Lip
  Generation In the Wild}. In \bibinfo{booktitle}{\emph{Proceedings of the 28th
  ACM International Conference on Multimedia}} \emph{(\bibinfo{series}{MM
  '20})}. \bibinfo{pages}{484–492}.
\newblock
\urldef\tempurl%
\url{https://doi.org/10.1145/3394171.3413532}
\showDOI{\tempurl}


\bibitem[Sajjadi et~al\mbox{.}(2018)]%
        {frvsr}
\bibfield{author}{\bibinfo{person}{Mehdi S.~M. Sajjadi},
  \bibinfo{person}{Raviteja Vemulapalli}, {and} \bibinfo{person}{Matthew
  Brown}.} \bibinfo{year}{2018}\natexlab{}.
\newblock \showarticletitle{{Frame-Recurrent Video Super-Resolution}}. In
  \bibinfo{booktitle}{\emph{{The IEEE Conference on Computer Vision and Pattern
  Recognition (CVPR)}}}.
\newblock


\bibitem[Siarohin et~al\mbox{.}(2019)]%
        {fomm_nips_2019}
\bibfield{author}{\bibinfo{person}{Aliaksandr Siarohin},
  \bibinfo{person}{Stéphane Lathuilière}, \bibinfo{person}{Sergey Tulyakov},
  \bibinfo{person}{Elisa Ricci}, {and} \bibinfo{person}{Nicu Sebe}.}
  \bibinfo{year}{2019}\natexlab{}.
\newblock \showarticletitle{First Order Motion Model for Image Animation}. In
  \bibinfo{booktitle}{\emph{Conference on Neural Information Processing Systems
  (NeurIPS)}}.
\newblock


\bibitem[Tang et~al\mbox{.}(2021)]%
        {depthSR}
\bibfield{author}{\bibinfo{person}{Jiaxiang Tang}, \bibinfo{person}{Xiaokang
  Chen}, {and} \bibinfo{person}{Gang Zeng}.} \bibinfo{year}{2021}\natexlab{}.
\newblock \bibinfo{booktitle}{\emph{Joint Implicit Image Function for Guided
  Depth Super-Resolution}}.
\newblock \bibinfo{publisher}{Association for Computing Machinery},
  \bibinfo{address}{New York, NY, USA}, \bibinfo{pages}{4390–4399}.
\newblock
\showISBNx{9781450386517}
\urldef\tempurl%
\url{https://doi.org/10.1145/3474085.3475584}
\showURL{%
\tempurl}


\bibitem[Thies et~al\mbox{.}(2018)]%
        {face2face}
\bibfield{author}{\bibinfo{person}{Justus Thies}, \bibinfo{person}{Michael
  Zollh\"{o}fer}, \bibinfo{person}{Marc Stamminger}, \bibinfo{person}{Christian
  Theobalt}, {and} \bibinfo{person}{Matthias Nie\ss{}ner}.}
  \bibinfo{year}{2018}\natexlab{}.
\newblock \showarticletitle{Face2Face: Real-Time Face Capture and Reenactment
  of RGB Videos}.
\newblock \bibinfo{journal}{\emph{Commun. ACM}} \bibinfo{volume}{62},
  \bibinfo{number}{1} (\bibinfo{date}{Dec.} \bibinfo{year}{2018}),
  \bibinfo{pages}{96–104}.
\newblock
\showISSN{0001-0782}
\urldef\tempurl%
\url{https://doi.org/10.1145/3292039}
\showDOI{\tempurl}


\bibitem[Timofte et~al\mbox{.}(2018)]%
        {ntire_report}
\bibfield{author}{\bibinfo{person}{Radu Timofte}, \bibinfo{person}{Shuhang Gu},
  \bibinfo{person}{Jiqing Wu}, {and} \bibinfo{person}{Luc Van~Gool}.}
  \bibinfo{year}{2018}\natexlab{}.
\newblock \showarticletitle{NTIRE 2018 Challenge on Single Image
  Super-Resolution: Methods and Results}. In
  \bibinfo{booktitle}{\emph{Proceedings of the IEEE Conference on Computer
  Vision and Pattern Recognition (CVPR) Workshops}}.
\newblock


\bibitem[Torralba et~al\mbox{.}(2007)]%
        {torralba_2006}
\bibfield{author}{\bibinfo{person}{Antonio Torralba}, \bibinfo{person}{Kevin~P.
  Murphy}, {and} \bibinfo{person}{William~T. Freeman}.}
  \bibinfo{year}{2007}\natexlab{}.
\newblock \showarticletitle{Sharing Visual Features for Multiclass and
  Multiview Object Detection}.
\newblock \bibinfo{journal}{\emph{IEEE Transactions on Pattern Analysis and
  Machine Intelligence}} \bibinfo{volume}{29}, \bibinfo{number}{5}
  (\bibinfo{year}{2007}), \bibinfo{pages}{854--869}.
\newblock
\urldef\tempurl%
\url{https://doi.org/10.1109/TPAMI.2007.1055}
\showDOI{\tempurl}


\bibitem[Wang et~al\mbox{.}(2019)]%
        {fsvid2vid_nips_2019}
\bibfield{author}{\bibinfo{person}{Ting-Chun Wang}, \bibinfo{person}{Ming-Yu
  Liu}, \bibinfo{person}{Andrew Tao}, \bibinfo{person}{Guilin Liu},
  \bibinfo{person}{Jan Kautz}, {and} \bibinfo{person}{Bryan Catanzaro}.}
  \bibinfo{year}{2019}\natexlab{}.
\newblock \showarticletitle{Few-shot Video-to-Video Synthesis}. In
  \bibinfo{booktitle}{\emph{Advances in Neural Information Processing Systems
  (NeurIPS)}}.
\newblock


\bibitem[Wang et~al\mbox{.}(2021)]%
        {nvidia_compression_2021_cvpr}
\bibfield{author}{\bibinfo{person}{Ting-Chun Wang}, \bibinfo{person}{Arun
  Mallya}, {and} \bibinfo{person}{Ming-Yu Liu}.}
  \bibinfo{year}{2021}\natexlab{}.
\newblock \showarticletitle{One-Shot Free-View Neural Talking-Head Synthesis
  for Video Conferencing}. In \bibinfo{booktitle}{\emph{Proceedings of the IEEE
  Conference on Computer Vision and Pattern Recognition}}.
\newblock


\bibitem[Wu et~al\mbox{.}(2021)]%
        {talkingfaceacmmm}
\bibfield{author}{\bibinfo{person}{Haozhe Wu}, \bibinfo{person}{Jia Jia},
  \bibinfo{person}{Haoyu Wang}, \bibinfo{person}{Yishun Dou},
  \bibinfo{person}{Chao Duan}, {and} \bibinfo{person}{Qingshan Deng}.}
  \bibinfo{year}{2021}\natexlab{}.
\newblock \bibinfo{booktitle}{\emph{Imitating Arbitrary Talking Style for
  Realistic Audio-Driven Talking Face Synthesis}}.
\newblock \bibinfo{publisher}{Association for Computing Machinery},
  \bibinfo{address}{New York, NY, USA}, \bibinfo{pages}{1478–1486}.
\newblock
\showISBNx{9781450386517}
\urldef\tempurl%
\url{https://doi.org/10.1145/3474085.3475280}
\showURL{%
\tempurl}


\bibitem[Xu et~al\mbox{.}(2014)]%
        {CompressionNN_nips_2014}
\bibfield{author}{\bibinfo{person}{Li Xu}, \bibinfo{person}{Jimmy~SJ Ren},
  \bibinfo{person}{Ce Liu}, {and} \bibinfo{person}{Jiaya Jia}.}
  \bibinfo{year}{2014}\natexlab{}.
\newblock \showarticletitle{Deep Convolutional Neural Network for Image
  Deconvolution}. In \bibinfo{booktitle}{\emph{Advances in Neural Information
  Processing Systems}}, Vol.~\bibinfo{volume}{27}. \bibinfo{publisher}{Curran
  Associates, Inc.}
\newblock


\bibitem[Yang(2019)]%
        {SRusingdual}
\bibfield{author}{\bibinfo{person}{Bin-Cheng Yang}.}
  \bibinfo{year}{2019}\natexlab{}.
\newblock \showarticletitle{Super Resolution Using Dual Path Connections}
  \emph{(\bibinfo{series}{MM '19})}. \bibinfo{publisher}{Association for
  Computing Machinery}, \bibinfo{address}{New York, NY, USA},
  \bibinfo{pages}{1552–1560}.
\newblock
\showISBNx{9781450368896}
\urldef\tempurl%
\url{https://doi.org/10.1145/3343031.3350878}
\showDOI{\tempurl}


\bibitem[Yao et~al\mbox{.}(2020)]%
        {Yao2020IterativeTE}
\bibfield{author}{\bibinfo{person}{Xin-Wei Yao}, \bibinfo{person}{Ohad Fried},
  \bibinfo{person}{K. Fatahalian}, {and} \bibinfo{person}{Maneesh Agrawala}.}
  \bibinfo{year}{2020}\natexlab{}.
\newblock \showarticletitle{Iterative Text-based Editing of Talking-heads Using
  Neural Retargeting}.
\newblock \bibinfo{journal}{\emph{ArXiv}}  \bibinfo{volume}{abs/2011.10688}
  (\bibinfo{year}{2020}).
\newblock


\bibitem[Zhang et~al\mbox{.}(2021)]%
        {facial_iccv_2021}
\bibfield{author}{\bibinfo{person}{Chenxu Zhang}, \bibinfo{person}{Yifan Zhao},
  \bibinfo{person}{Yifei Huang}, \bibinfo{person}{Ming Zeng},
  \bibinfo{person}{Saifeng Ni}, \bibinfo{person}{Madhukar Budagavi}, {and}
  \bibinfo{person}{Xiaohu Guo}.} \bibinfo{year}{2021}\natexlab{}.
\newblock \showarticletitle{FACIAL: Synthesizing Dynamic Talking Face with
  Implicit Attribute Learning}. In \bibinfo{booktitle}{\emph{Proceedings of the
  IEEE/CVF International Conference on Computer Vision (ICCV)}}.
  \bibinfo{pages}{3867--3876}.
\newblock


\bibitem[Zhang et~al\mbox{.}(2017)]%
        {S3FDFaceDet_iccv_2017}
\bibfield{author}{\bibinfo{person}{Shifeng Zhang}, \bibinfo{person}{Xiangyu
  Zhu}, \bibinfo{person}{Zhen Lei}, \bibinfo{person}{Hailin Shi},
  \bibinfo{person}{Xiaobo Wang}, {and} \bibinfo{person}{S. Li}.}
  \bibinfo{year}{2017}\natexlab{}.
\newblock \showarticletitle{S${^3}$FD: Single Shot Scale-Invariant Face
  Detector}.
\newblock \bibinfo{journal}{\emph{2017 IEEE International Conference on
  Computer Vision (ICCV)}} (\bibinfo{year}{2017}), \bibinfo{pages}{192--201}.
\newblock


\bibitem[Zhou et~al\mbox{.}(2021)]%
        {PC_AVS_2021_cvpr}
\bibfield{author}{\bibinfo{person}{Hang Zhou}, \bibinfo{person}{Yasheng Sun},
  \bibinfo{person}{Wayne Wu}, \bibinfo{person}{Chen~Change Loy},
  \bibinfo{person}{Xiaogang Wang}, {and} \bibinfo{person}{Ziwei Liu}.}
  \bibinfo{year}{2021}\natexlab{}.
\newblock \showarticletitle{Pose-Controllable Talking Face Generation by
  Implicitly Modularized Audio-Visual Representation}. In
  \bibinfo{booktitle}{\emph{Proceedings of the IEEE Conference on Computer
  Vision and Pattern Recognition (CVPR)}}.
\newblock


\bibitem[Zhou et~al\mbox{.}(2020)]%
        {Yang:2020:MakeItTalk}
\bibfield{author}{\bibinfo{person}{Yang Zhou}, \bibinfo{person}{Xintong Han},
  \bibinfo{person}{Eli Shechtman}, \bibinfo{person}{Jose Echevarria},
  \bibinfo{person}{Evangelos Kalogerakis}, {and} \bibinfo{person}{Dingzeyu
  Li}.} \bibinfo{year}{2020}\natexlab{}.
\newblock \showarticletitle{MakeItTalk: Speaker-Aware Talking-Head Animation}.
\newblock \bibinfo{journal}{\emph{ACM Transactions on Graphics}}
  \bibinfo{volume}{39}, \bibinfo{number}{6} (\bibinfo{year}{2020}).
\newblock


\end{thebibliography}

\newpage
\appendix

\section{SUPPLEMENTARY MATERIAL}

\subsection{Additional Network and Training Details}

\subsubsection{Frame-Interpolation Network}

\begin{figure}[ht]
\centering
  \includegraphics[width=\linewidth]{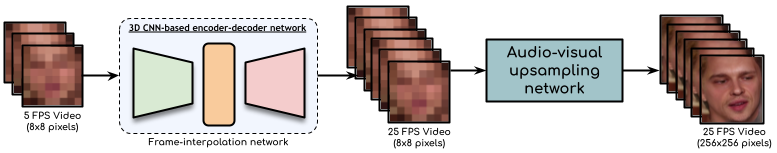}
  \vspace{-20pt}
  \caption{Our designed pipeline to transmit $5$FPS LR videos and generate the final $25$FPS HR videos. We achieve a major reduction in the bits/pixel (BPP) using our frame-interpolation network, since only $1$ in $5$ LR frames needs to be transmitted.}
  \label{fig:frame_interpolation}
  \vspace{-10pt}
\end{figure}

As discussed in the paper, we design a frame-interpolation network to further reduce the bandwidth consumption. Our frame-interpolation network is a simple encoder-decoder based architecture, majorly consisting of $3$D CNN layers. The pipeline for transmitting $5$FPS low-resolution (LR) frames and applying our frame-interpolation network to generate $25$FPS LR frames, and further upsampling them to obtain the high-resolution (HR) talking-face video as output is shown in Figure~\ref{fig:frame_interpolation}. 

\subsubsection{Training Details}
For training, we randomly sample a contiguous window of $T=5$ ($0.2$ seconds) frames from the AVSpeech~\cite{avspeech_2018_tog} or VoxCeleb2~\cite{voxceleb2_2018_interspeech} datasets, resized to $256\times256$ pixels. We extract the corresponding $0.2$ seconds audio signal sampled at $16$kHz. Bicubic degradation is used to downscale the ground-truth images $F_{gt}$ and generate the corresponding LR images $F_{lr}$ at the given scale-factor. We train our network using the Adam optimizer~\cite{adam} with an initial learning rate of $1e^{-4}$, with a learning rate decay by a factor of $10$ after every half of the total epochs. The training is stopped once the validation loss does not improve for $10$ epochs.

\subsubsection{Metrics}
We use multiple metrics to evaluate the different aspects of the results. The details of these metrics are given below.

\noindent
\textbf{PSNR:} PSNR measures the mean squared error (MSE) between the generated and ground-truth frames, thus correlating with the reconstruction quality. 

\noindent
\textbf{SSIM:} To measure the structural similarity between the patches of the frames, we use the SSIM metric. SSIM is more robust to global illumination changes than PSNR, which only considers pixel-wise absolute errors.

\noindent
\textbf{FID:} To evaluate the perceptual quality of the generated frames, we use Fréchet Inception Distance (FID), which measures the distance between the generated and the ground-truth distributions.

\noindent
\textbf{LMD:} To evaluate the structure of the generated faces, we report the Landmark-distance (LMD) using FAN-based~\cite{fan_landmark_iccv_2017} landmark detector. We compute the landmark distance for face regions like eyes, lips, eyebrows and nose for the generated and ground-truth frames. 

\noindent
\textbf{LSE-D:} We follow Wav2Lip~\cite{wav2lip} in measuring the lip-sync accuracy by using the Lip Sync Error Distance (LSE-D) metric calculated using the pre-trained SyncNet\footnote{\url{https://github.com/joonson/syncnet_python}}~\cite{syncnet_2016}.

\subsection{Additional Experiments and Evaluations}

\subsubsection{SR at multiple scale-factors}

\begin{figure*}[ht]
\centering
  \includegraphics[width=\linewidth]{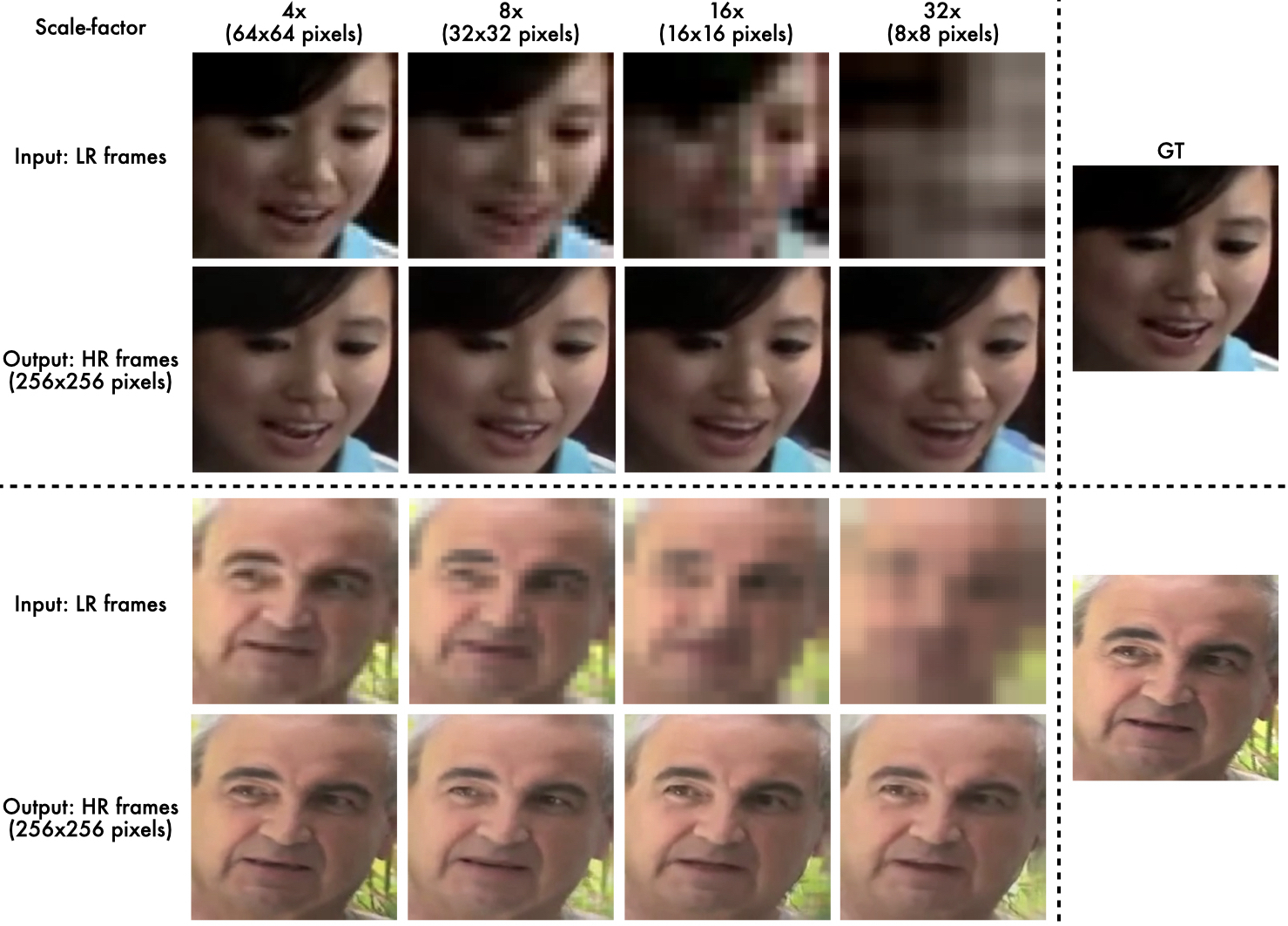}
  \vspace{-20pt}
  \caption{Qualitative results of our method at different scale-factors ($4\times$, $8\times$, $16\times$ and $32\times$). We can observe that our approach generates consistent results across all scales. It is impressive to note the high-quality results at high scale-factors like $32\times$.}
  \label{fig:scale_images}
\end{figure*}

\begin{figure}[ht]
\centering
  \includegraphics[width=\linewidth]{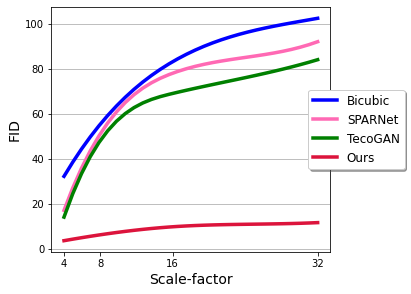}
   \vspace{-20pt}
  \caption{Performance comparison of different methods (metric: FID) at multiple scale-factors. The gap in the performance of our method versus the existing approaches, especially at higher scale-factors like $32\times$ indicates the capability and  robustness of our technique.}
  \label{fig:scale}
\end{figure}

To analyze the performance of our model at various scale-factors, we plot the variation of the FID metric across the scale-factors of $4\times$, $8\times$, $16\times$, and $32\times$ in Figure~\ref{fig:scale}. We can observe that the gap in performance of our method versus the existing approaches increases as the scale-factor increases. At smaller scale-factors like $4\times$, the existing models SPARNet~\cite{SPARNet} and TecoGAN~\cite{tecogan} performs decently, however, at higher scale-factors, our model generates significantly better results. This demonstrates that our model is able to recover the lost details and produces plausible high-quality results even from very low-resolution inputs.

Figure~\ref{fig:scale_images} shows the qualitative results generated by our upsampling models at multiple scale- factors. We can observe consistent performance of our models across the scale-factors. The level of facial details captured by our method is impressive, even at higher scale-factors like $32\times$. The lip shape, hair, and teeth (Figure~\ref{fig:scale_images} - row $1$) and eyebrows, face texture and eyeballs (Figure~\ref{fig:scale_images} - row $2$) demonstrates that our model generates sharp, high-frequency details, thus preserving most of the identity-specific information.

\subsubsection{Analysis of the Target Identity Image}
Our network takes a single HR target identity image as input. As discussed in the paper, this image can be any sample frame, either from the same video or any other image, with similar characteristics in terms of the face identity, pose, clothing and background. We empirically verify this below using two cases, one where we provide a target image of the same identity but from a different video, and another where we provide a different identity altogether.

\begin{figure}[ht]
\centering
  \includegraphics[width=\linewidth]{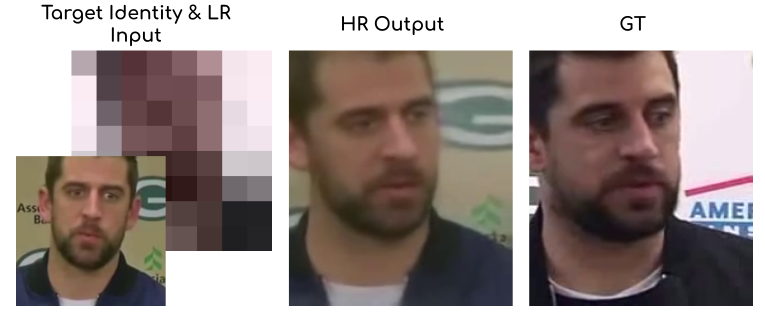}
  \vspace{-20pt}
  \caption{Our method works decently when the target identity image given as input is from a different video of the same person.}
  \label{fig:same_id_diff_ref}
   \vspace{-10pt}
\end{figure}

\paragraph{(i) Same Identity - Different Image:}
As shown in Figure~\ref{fig:same_id_diff_ref}, we can see that our model is able to generate decent results for an input target identity image sampled from a different video, with slight variations in terms of clothing, illumination, background and pose. Although there is room for improvement in quality, we can see that our model captures the motion well and is also able to replicate the basic identity details.

\paragraph{(ii) Different Identity Image:}

\begin{figure}[ht]
\centering
  \includegraphics[width=\linewidth]{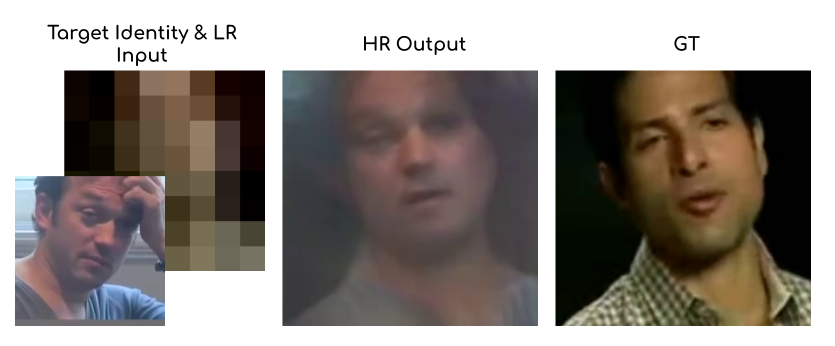}
  \vspace{-20pt}
  \caption{Our method does not generate intelligible HR outputs if the target identity image given as input is from a different person than the LR input video, thus avoiding direct creation of ``deepfakes''.}
  \label{fig:diff_id_diff_ref}
   \vspace{-10pt}
\end{figure}

Figure~\ref{fig:diff_id_diff_ref} depicts our results when the input identity image is from a different identity altogether. As Section $5$ of the paper explains, our model does not work if the target identity and the LR frames do not match. This is one of the advantages of our method since the creation of harmful ``deepfakes'' is directly not possible from our model. We encourage the readers to see our demo video for the video results.

\subsubsection{Human Evaluations}

\begin{figure}[t]
  \includegraphics[width=\linewidth]{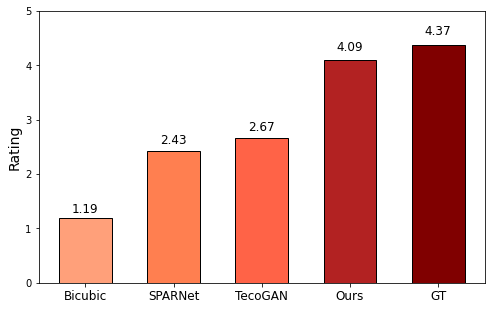}
  \vspace{-20pt}
  \caption{Mean opinion scores of various methods obtained from human evaluations. Higher scores indicate better quality and perceptual satisfaction.}
  \label{fig:human_eval}
\end{figure}

We conduct human evaluations as a part of subjective assessment, primarily to evaluate the quality of the generated videos. We randomly select $20$ videos from the curated AVSpeech test set~\cite{avspeech_2018_tog} and in each video, we select a random contiguous $5$ seconds video segment (a window of $125$ frames). The results for these segments generated from all the comparison methods, our approach and the corresponding ground-truth videos are displayed in random order. We ask $30$ participants to rate these videos on a scale of $1-5$ based on the perceptual satisfaction. The participant group spans members of $22$ - $40$ years with an almost equal male-female ratio. The users are asked to consider the overall quality based on the facial texture, lip-sync and the sharpness of frames. The mean opinion scores of all the participants are displayed in Figure~\ref{fig:human_eval}. In-line with our qualitative and quantitative results, our approach outperforms all the other methods by a very large margin. Another interesting point to note is that our scores are quite close to the scores of ground-truth videos, indicating that we are able to recover the actual details to a large extent.

\subsection{Additional Qualitative Results}

\subsubsection{Talking-Face Video Upsampling}
\begin{figure*}[ht]
\centering
  \includegraphics[width=\linewidth]{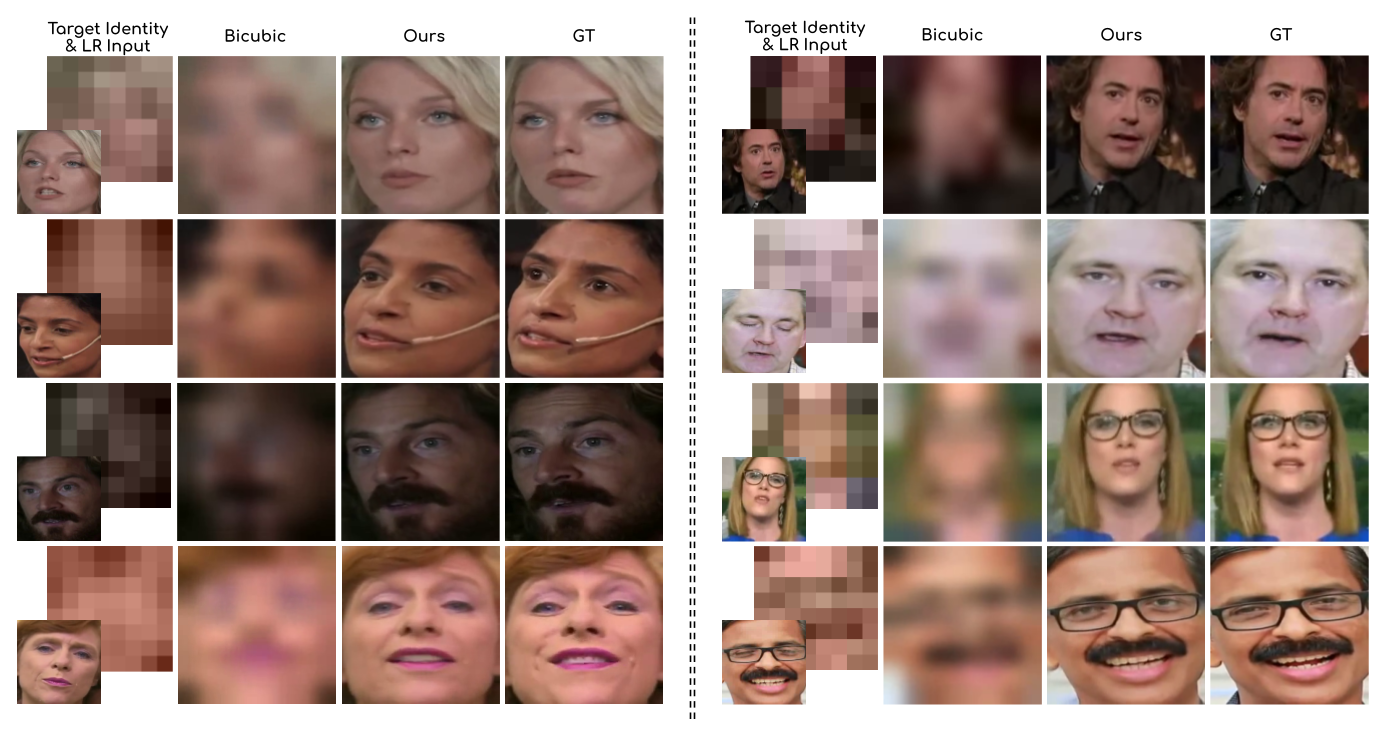}
  \vspace{-20pt}
  \caption{Additional qualitative results: Our model captures the identity-specific facial details and generates faithful results.}
  \label{fig:more_results}
\end{figure*}

We show additional qualitative samples generated by our model in Figure~\ref{fig:more_results}. In-line with the previous results, our model is able to accurately match the ground-truth and replicate the identity-specific details very well.

\begin{figure*}[ht]
\centering
  \includegraphics[width=\linewidth]{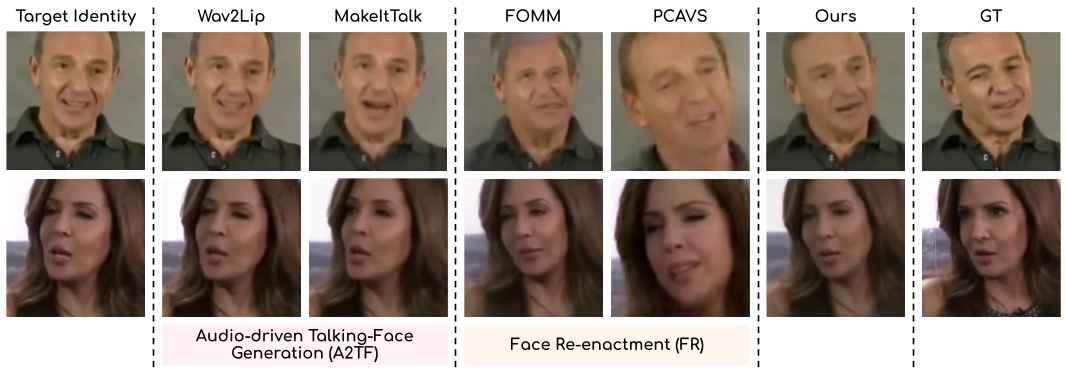}
   \vspace{-20pt}
  \caption{Qualitative comparison with Audio-driven Talking-Face Generation (A2TF) and Face Re-enactment (FR) methods. A2TF methods fail to preserve the accurate pose of the actual identity; FR methods also face similar issues and also degrades the quality of generations. In contrast, our method produces the best results in comparison, preserves the actual identity and also generates high-quality outputs.}
  \label{fig:a2tf_fr_comparison}
\end{figure*}

\subsubsection{Audio-driven Talking-Face Generation (A2TF) and Face Re-enactment (FR) comparison}

In Section $4.3$ of our main paper, we demonstrated the quantitative comparisons with A2TF and FR methods. To further understand and analyse these comparisons, we include qualitative comparison in Figure~\ref{fig:a2tf_fr_comparison}. Similar to Table $4$, we compare with two A2Tf works, (i) Wav2Lip~\cite{wav2lip} and (ii) MakeItTalk~\cite{Yang:2020:MakeItTalk} and two FR works, (i) FOMM~\cite{fomm_nips_2019} and (ii) PCAVS~\cite{pcavs}. From the figure, we can infer that the A2TF methods do not preserve the head movements of the original video. This is because these methods ingest a single target identity image and animate only the lip region (lower-half of the face), thus the upper half of the face, i.e., the head pose remains static (as that of the target identity input). The results generated by FR methods are also not very accurate, in terms of the pose and also the generation quality. In comparison, we find that our method produces the best results, thus validating the robustness of our approach. Please refer to our demo video for full video comparison.

\subsection{Ablation Experiments}
In order to analyse the various design choices of our network, we perform several ablation experiments. All the results are reported on the test set of AVSpeech dataset~\cite{avspeech_2018_tog}.

\subsubsection{Importance of the Audio Signal}

We train a network by discarding the audio input in our proposed network to highlight the importance of the audio signal. As shown in Table~\ref{table:ablation_audio}, it results in poor reconstruction of faces and inaccurate lip shape generation (indicated by LSE-D metric). This signifies the importance of the audio information in generating plausible faces as well as precise lip and mouth shapes. Without the audio signal, the model does not have adequate information in the LR input to recover the fine-grained lip shape, an essential feature in talking-face videos. The audio information also assists the other facial features that define the identity and appearance of a person. 

\begin{figure}[t]
  \includegraphics[width=\linewidth]{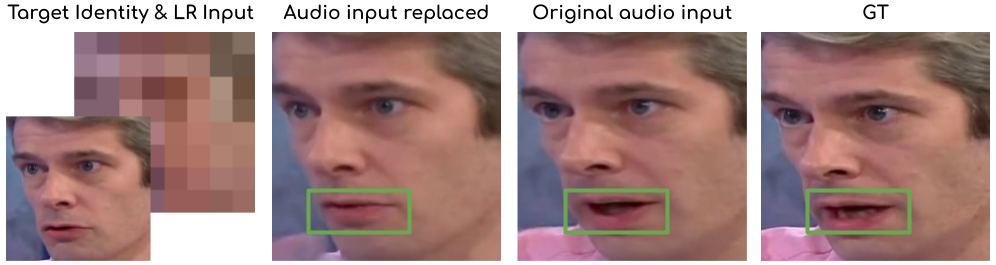}
  \vspace{-20pt}
  \caption{Audio signal helps in accurate lip-shape generation.}
  \label{fig:replace_audio}
  \vspace{-10pt}
\end{figure}

We also show a qualitative sample in Figure~\ref{fig:replace_audio} to illustrate the benefits of incorporating the audio signal in our pipeline. We replace the audio input with an audio signal from a different speaker. From Figure~\ref{fig:replace_audio}, we can observe that the audio signal is crucial for precise lip-shape generation and identity recovery.

\begin{table}[ht]
    \centering

    \small
    \setlength{\tabcolsep}{4pt}
    \caption{The audio signal helps to recover basic identity details and also enforces accurate lip shape generation.}
    \vspace{-10pt}
    \begin{tabular}{c|cccccc}
    \hline
    
    \textbf{Method} & 
    PSNR$\uparrow$ & SSIM$\uparrow$ & FID$\downarrow$ & LMD$\downarrow$ & LSE-D$\downarrow$  \\
    
    \hline
    
    W/o audio & 24.09 & 0.70 & 15.14 & 0.178 & 13.98 \\
    
    With audio & 25.06 & 0.73 & 11.54 & 0.162 & 12.43\\

    \hline
    
    \end{tabular}
    \vspace{-10pt}
    \label{table:ablation_audio}
\end{table}

\subsubsection{Significance of our Proposed Landmark-based Region Loss}
\begin{figure}[ht]
  \includegraphics[width=\linewidth]{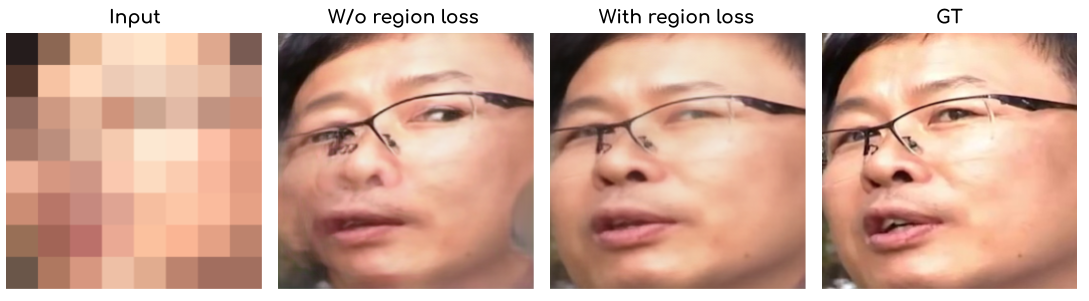}
  \vspace{-20pt}
  \caption{Optimizing the landmark-based region loss improves the overall generation quality by handling the pose variations and predicting the accurate face regions.}
  \label{fig:ablation_landmark}
\end{figure}

We assess the importance of optimizing the landmark-based region loss in Table~\ref{table:ablation_landmark} and Figure~\ref{fig:ablation_landmark}. Without using the region loss, the model struggles to estimate the accurate pose of different face regions, for example, eyes and nose (see Figure~\ref{fig:ablation_landmark} - column $2$). A large difference, especially in the LMD metric, can be observed in Table~\ref{table:ablation_landmark}. This occurs due to the lack of supervision regarding the human face structure. We provide this additional control using the landmark-based region loss, which helps not only in enforcing the local correspondence, but also assists in smooth transitions for pose variations (can be seen in the demo video).

\begin{table}[ht]
    \centering

    \small
    \setlength{\tabcolsep}{4pt}
    \caption{The landmark-based region loss helps to preserve the overall face structure.}
    \vspace{-10pt}
    \begin{tabular}{c|cccccc}
    \hline
    
    \textbf{Method} & 
    PSNR$\uparrow$ & SSIM$\uparrow$ & FID$\downarrow$ & LMD$\downarrow$ & LSE-D$\downarrow$  \\
    
    \hline
    
    W/o region loss & 23.96 & 0.67 & 15.54 & 0.230 & 13.12\\
    
    With region loss & 25.06 & 0.73 & 11.54 & 0.162 & 12.43\\

    \hline
    
    \end{tabular}
    \vspace{-10pt}
    \label{table:ablation_landmark}
\end{table}

\subsubsection{Importance of the Backbone Network}

As discussed in Section $3.2$ of the paper, we pre-train our backbone network using the input LR frames and audio signal, to extract the elementary facial details and motion information. To analyze the importance of pre-training, we directly train our animation model end-to-end without pre-training the backbone. Table~\ref{table:ablation_backbone} shows the results. We can notice the effect of pre-training in Table~\ref{table:ablation_backbone}. The outputs from the backbone network are considered as ``driving video'' in our end-to-end trainable animation model; thus, capturing the basic face details is crucial for the model to work well.

\begin{table}[ht]
    \centering

    \small
    \setlength{\tabcolsep}{4pt}
    \caption{Pre-training the backbone network helps to capture the basic face details and motion information, thus improving the overall model's performance.}
    \vspace{-10pt}
    \begin{tabular}{c|cccccc}
    \hline
    
    \textbf{Method} & 
    PSNR$\uparrow$ & SSIM$\uparrow$ & FID$\downarrow$ & LMD$\downarrow$ & LSE-D$\downarrow$  \\
    
    \hline
    
    W/o pre-training & 21.90 & 0.69 & 19.23 & 0.220 & 16.61 \\
    
    With pre-training & 25.06 & 0.73 & 11.54 & 0.162 & 12.43\\

    \hline
    
    \end{tabular}
    \label{table:ablation_backbone}
\end{table}

\subsubsection{Temporal Window}
We justify the use of $T=5$ contiguous frames in our network in Table~\ref{table:ablation_frames}. The performance of the network deteriorates when only $3$ frames are used. This is mainly because the network does not learn the temporal aspect without adequate context. The performance of using $5$ and $7$ frames are very similar, which suggests that the context of $5$ frames is sufficient for the model, especially to learn specialized temporal aspects such as lip-sync. This is inline with the previous work~\cite{wav2lip} where it is shown that the best performance is achieved using a window of $5$ frames.

\begin{table}[ht]
    \centering

    \small
    \setlength{\tabcolsep}{6pt}
    \caption{A context window of $T=5$ contiguous frames allows the model to effectively capture the temporal relationships such as lip-sync.}
    \vspace{-10pt}
    \begin{tabular}{c|ccccc}
    \hline
    
    \textbf{Frames} & 
    PSNR$\uparrow$ & SSIM$\uparrow$ & FID$\downarrow$ & LMD$\downarrow$ & LSE-D$\downarrow$  \\
     
    \hline
    
    $3$ & 23.88 & 0.69 & 15.78 & 0.187 & 12.78\\
    
    $5$ & 25.06 & 0.73 & 11.54 & 0.162 & 12.43\\

    $7$ & 25.13 & 0.74 & 11.10 & 0.157 & 12.28\\
    
    \hline
    
    \end{tabular}
    \label{table:ablation_frames}
\end{table}

\subsection{A Baseline Model for Handling Emotions}

\begin{figure*}[ht]
\centering
  \includegraphics[width=\linewidth]{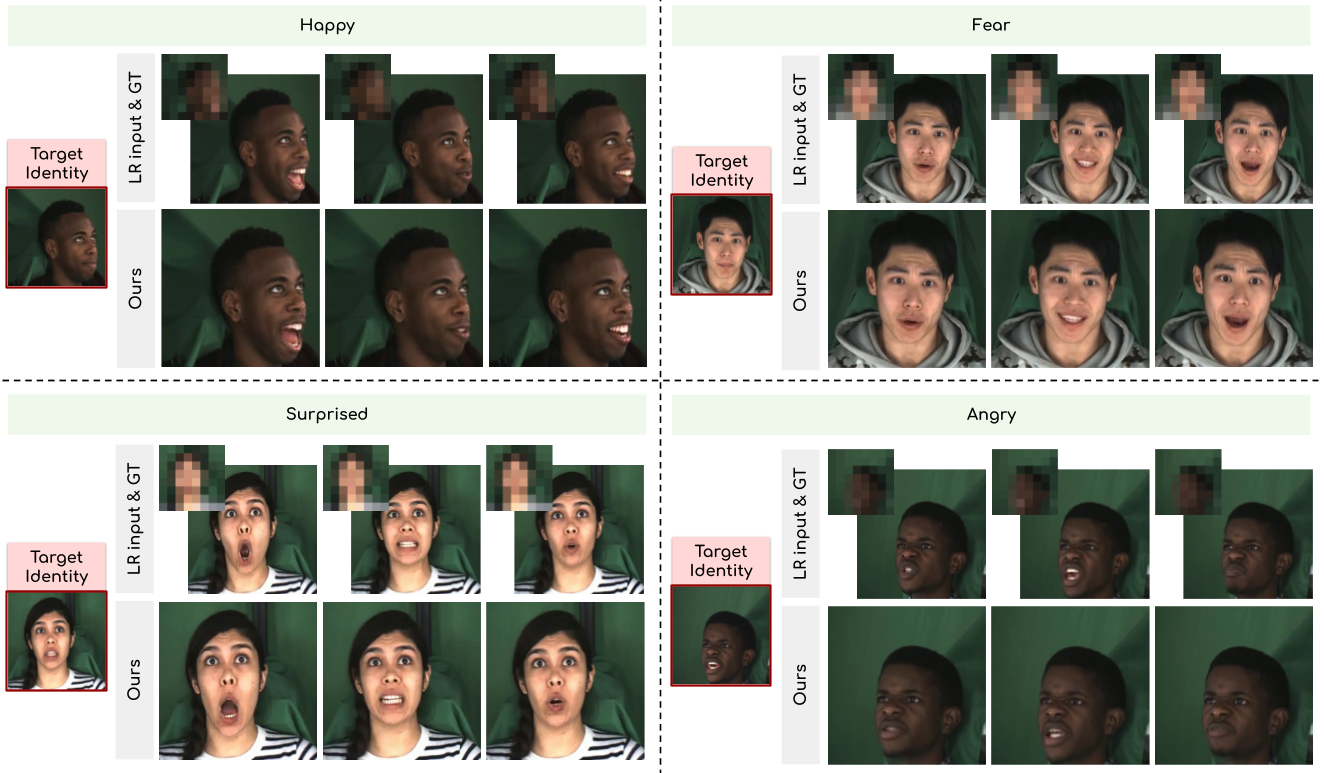}
  \vspace{-20pt}
  \caption{Qualitative results from our model on the emotional dataset, MEAD~\cite{MEAD_eccv_2020}. Our model handles the emotions and generates consistent results across all the emotions.}
  \label{fig:mead_results}
\end{figure*}

The datasets we considered so far, namely, AVSpeech~\cite{avspeech_2018_tog} and VoxCeleb2~\cite{voxceleb2_2018_interspeech} do not contain emotion annotations and are not specifically designed to evaluate the expressions. Thus, we consider to use the large-scale emotional dataset, ``Multi-view Emotional Audio-visual Dataset'' (MEAD)~\cite{MEAD_eccv_2020}. MEAD is a challenging corpus comprising eight different emotions from $60$ actors at different intensity levels, captured at multiple view angles. To validate the capability of our model in handling the variations in expressions, we train our talking-face upsampling model on MEAD. Note that we do not add or modify our architecture for this training, all the training settings are the same as we used for other datasets.

Figure~\ref{fig:mead_results} shows the results generated by our model. We can see that our model is able to capture emotions very well. Since MEAD contains large amount of speaker-specific data, our model learns the fine-grained characteristics of individual speakers. This is reflected in the results shown in Figure~\ref{fig:mead_results}. Details like wide opening of mouth, glabellar lines (wrinkles) on the forehead and dimples near teeth are also synthesized by our model. Appropriate extensions (like emotion classifier discriminator) can be added to our model in the future to be able to handle extreme variations of expressions (like shown in Figure $8 (b)$ of the main paper).

\end{document}